\documentclass[letterpaper, 10 pt, conference]{ieeeconf}  % Comment this line out if you need a4paper

\IEEEoverridecommandlockouts                              

\let\labelindent\relax

\usepackage{enumitem}
\usepackage{amsthm}
\usepackage[T1]{fontenc}
\usepackage{xcolor}
\usepackage{amsmath}
\usepackage{amssymb}
\usepackage{algorithm}
\usepackage{algpseudocode}
\usepackage{graphicx} 
\usepackage{subcaption}
\usepackage{float}
\usepackage{placeins}
\usepackage{cleveref}

\setlist[itemize]{leftmargin=1.25em, itemsep=2pt}
\setlist[enumerate]{leftmargin=1.25em, itemsep=2pt}

\theoremstyle{plain}
\newtheorem{proposition}{Proposition}

\theoremstyle{definition}
\newtheorem{definition}{Definition}
\newtheorem{example}{Example}
\newtheorem{remark}{Remark}
\newtheorem{assumption}{Assumption}

\title{\LARGE \bf
Transformer-Based Multi-Agent Reinforcement Learning for Networked Systems with Long-Range Interactions
}

\author{
Vidur Sinha$^{1}$, Muhammed Ustaomeroglu$^{1}$, Guannan Qu$^{1}$%
\thanks{$^{1}$Vidur Sinha, Muhammed Ustaomeroglu, and Guannan Qu are with the Department of Electrical and Computer Engineering,
Carnegie Mellon University, Pittsburgh, USA. Correspondence to: Vidur Sinha <vidursin@andrew.cmu.edu>.}
}

\begin{document}

\maketitle
\thispagestyle{empty}
\pagestyle{empty}

%%%%%%%%%%%%%%%%%%%%%%%%%%%%%%%%%%%%%%%%%%%%%%%%%%%%%%%%%%%%%%%%%%%%%%%%%%%%%%%%
\begin{abstract}
Multi-agent reinforcement learning (MARL) has shown promise for large-scale network control, yet existing methods face two major limitations. First, they typically rely on an exponential decay property of agent interactions on far-away nodes, which can be exploited to develop more efficient and tractable MARL algorithms. When this exponential decay property does not hold, these algorithms do not account for \emph{long-range interactions} such as epidemic outbreaks or cascading power failures. Second, existing approaches lack \emph{network generalizability}, or the ability to generalize to networks of different topological structure and scale than those seen during training. In this work, we first present a mean-field stability analysis and empirical study investigating the conditions for long-range network interactions. These results motivate our primary contribution: STACCA (Shared Transformer Actor-Critic with Counterfactual Advantage), a transformer-based MARL framework that addresses both long-range interactions and network generalizability. STACCA employs a centralized Graph Transformer Critic to model long-range dependencies and provide system-level feedback, while its shared Graph Transformer Actor learns a generalizable policy capable of adapting across diverse network topologies. To improve credit assignment during training, STACCA integrates a novel counterfactual advantage estimator that is compatible with state-value critic estimates. We evaluate STACCA on epidemic containment and rumor-spreading network control tasks, demonstrating improved performance and network generalizability. These results highlight the potential of transformer-based MARL architectures to achieve generalizable control in large-scale networked systems.
\end{abstract}

%%%%%%%%%%%%%%%%%%%%%%%%%%%%%%%%%%%%%%%%%%%%%%%%%%%%%%%%%%%%%%%%%%%%%%%%%%%%%%%%
\section{Introduction}
\label{intro}

Multi-agent networked systems are ubiquitous in modern infrastructure, spanning power grids, transportation networks, social platforms, and beyond \cite{herrera2020multi, chen2022reinforcement}. These systems involve a large number of interacting agents whose states and actions are coupled by the underlying network topology and dynamics. While many networked systems exhibit a local interaction structure, in which agents can only immediately impact their local neighborhood (e.g., diseases spreading through contact), these local changes can sometimes trigger spatial or temporal long-range effects, such as epidemic outbreaks or cascading power failures. Thus, designing local decision policies that capture complex long-range dependencies among agents is essential for effective control in many large-scale networked systems.

Multi-agent reinforcement learning (MARL) has emerged as a powerful paradigm for such networked control problems. There are many existing general-purpose MARL algorithms (e.g., MADDPG \cite{Lowe2017MAACMixedEnvironments}, MAPPO \cite{Yu2022SurprisingPPO}); however, these methods are known to suffer from scalability issues, as the state and action spaces are exponentially large with respect to the number of agents. To address this issue, existing approaches have identified and exploited an exponential decay property in many networked MARL problems, which holds under certain ergodicity assumptions \cite{qu2020scalable_local, qu2020scalable}. This property indicates that an agent's state has negligible impact on far-away nodes (measured by graph distance), and it allows for tractable approximations of the Q-function through \emph{truncated} Q-functions which depend on a smaller spatial horizon. While these approaches allow for a rich class of networked MARL problems to be solved in a scalable manner (e.g., decentralized voltage control \cite{Xu2023ScalableNetworkAware}), two key challenges remain in networked multi-agent systems:

\textbf{(1) Long-Range Interactions.} A networked system has long-range interactions if a local state change (e.g., a person becoming infected or a tripped power line) can have non-local effects (e.g., an epidemic or cascading power failure). Existing approaches focus on the case in which long-range interactions have a negligible impact on the system (i.e., cases in which the exponential decay property holds), but there are many cases in which they have catastrophic impact. Thus, it is important to study the conditions under which long-range interactions occur and to develop policies that can effectively handle them. 

\textbf{(2) Network Generalizability.} Existing networked MARL solutions are typically trained and evaluated on fixed networks, with limited ability to generalize to unseen topologies. Network generalizability is a stronger condition than scalability, as it is the ability for a control policy to generalize not only to networks of large scale, but also of diverse structure. However, decentralized, network-generalizable policies are especially difficult to achieve in real-world networks where local neighborhoods are heterogeneous. Furthermore, the large number of agents required to train them exacerbates the infamous credit-assignment problem in MARL \cite{chang2003all}, making learning such policies incredibly challenging.

These challenges motivate the following questions: 
\begin{itemize}
    \item \textbf{Question 1: } \emph{Can we better understand long-range interactions in networked systems?}
    \item \textbf{Question 2: } \emph{Can we create a MARL solution for networked systems that handles long-range interactions and is network generalizable?}
\end{itemize}

\textbf{Contributions.} We approach Question 1 from a stability perspective and analyze the linearized mean-field system around a \emph{normal equilibrium} of the dynamics (e.g., disease-free state). We then connect this mean-field stability analysis to long-range interactions under the true system dynamics through an empirical study. Our results show that the spectral properties of the linearized mean-field dynamics reveal strong predictors for long-range interactions. These results motivate our solution to Question 2.

To address Question 2, we introduce STACCA (Shared Transformer Actor-Critic with Counterfactual Advantage), a transformer-based MARL framework. Transformers are known for their ability to model long-range dependencies, and their self-attention mechanism has been adapted to many domains, including graphs \cite{Vaswani2017AttentionAllYouNeed, velickovic2018graph}. STACCA employs a \textbf{(1) Centralized Graph Transformer Critic} that learns long-range dependencies among agents, providing global context to aid policy learning and enhance local decisions. Complementing this, a \textbf{(2) Shared Graph Transformer Actor} learns a policy that effectively \emph{stacca} (``detaches'' in Italian) from the specific network seen during training, enabling zero-shot transfer to unseen network topologies and scales. To both encourage the critic to learn long-range dependencies and expose the actor to diverse local network structures, we must train on relatively large networks, which amplifies the challenge of \emph{multi-agent credit assignment}. To address this, STACCA integrates a novel \textbf{(3) Counterfactual Advantage} that leverages the global state-value critic (as opposed to a less stable action-value critic) to isolate each agent’s contribution against a counterfactual baseline (i.e., what would have happened if the agent had acted differently).

Our experiments on epidemic containment and rumor-spreading network control tasks demonstrate that STACCA substantially outperforms the baseline MAPPO algorithm and all ablation variants, while also exhibiting zero-shot generalization across network topologies. These results highlight the potential of transformer-based MARL architectures to achieve scalable, transferable, and effective control in complex, large-scale networked systems.

\section{Problem Formulation and Background}
\label{problem_formulation}
We formulate the problem as a networked Markov Decision Process (MDP) and use it to introduce two network control examples: epidemic containment and rumor spreading. We then review key ideas regarding Multi-Agent Proximal Policy Optimization (MAPPO) \cite{Yu2022SurprisingPPO}, the Transformer \cite{Vaswani2017AttentionAllYouNeed}, and Graph Attention Networks (GATs) \cite{velickovic2018graph}.

\subsection{Problem Formulation}
We adopt a networked MDP framework based on \cite{qu2020scalable_local}. Consider an undirected graph $\mathcal{G} = (\mathcal{N}, \mathcal{E})$, where nodes $\mathcal{N}=\{1,\ldots,N\}$ are agents and $\mathcal{E}$ are the edges. The global state at time $t$ is the collection of individual agent states, $s_t = (s_{1,t}, \dots, s_{N,t})\in \mathcal{S} = \mathcal{S}_1\times\cdots\times \mathcal{S}_N$, with $s_{i,t}\in \mathcal{S}_i$. Similarly, the global action is $a_t = (a_{1,t},\ldots,a_{N,t})\in \mathcal{A} = \mathcal{A}_1\times \cdots \times \mathcal{A}_N$, and the global observation is $o_t = (o_{1,t},\ldots,o_{N,t})$. Each agent's local observation is given by the states of the agents in its closed $k$-hop neighborhood $\mathcal{N}_i^k$ (includes node $i$) and the edges of the corresponding subgraph: $o_{i,t} = (s_{\mathcal{N}_i^k,t}, \mathcal{E}_{\mathcal{N}_i^k})$.

A key property of our model is that state transitions factorize across each agent's closed 1-hop neighborhood:\footnote{We exclude the superscript for the closed $k$-hop neighborhood when $k$=1.}
\begin{equation} 
\label{eq:factorize}
P(s_{t+1}|s_t,a_t) = \prod_{i=1}^{N} P_i(s_{i,t+1} | s_{\mathcal{N}_i,t}, a_{\mathcal{N}_i,t})
\end{equation}
Further, we let $r(s,a)$ denote a shared reward among all agents and $\gamma \in [0, 1)$ be the discount factor. In MARL, typically each agent acts according to a memoryless decentralized policy $\pi_i(a_i|o_i)$, using only its local observation $o_i$. The objective is to maximize the expected team return
\begin{equation}
J(\boldsymbol{\pi}) = \mathbb{E}_{\boldsymbol{\pi}}\!\left[ \sum_{t=0}^{\infty}\gamma^t r(s_t, a_t) \right],
\end{equation}
where $\boldsymbol{\pi}=(\pi_1,\dots,\pi_N)$. Correspondingly, the global state-value function under policy $\boldsymbol{\pi}$ is defined as
\begin{equation}
V_{\boldsymbol{\pi}}(s_t) = \mathbb{E}_{\boldsymbol{\pi}}\Bigg[\sum_{k=0}^{\infty} \gamma^k r(s_{t+k}, a_{t+k}) \,\Big|\, s_t \Bigg],
\label{eq:value_function}
\end{equation}
and the advantage of taking joint action $a_t$ in state $s_t$ is
\begin{equation}
A_{\boldsymbol{\pi}}(s_t, a_t) = Q_{\boldsymbol{\pi}}(s_t, a_t) - V_{\boldsymbol{\pi}}(s_t),
\label{eq:advantage_function}
\end{equation}
where $Q_{\boldsymbol{\pi}}(s_t, a_t) = \mathbb{E}_{\boldsymbol{\pi}}[\sum_{k=0}^{\infty} \gamma^k r(s_{t+k}, a_{t+k}) | s_t, a_t]$ is the joint action-value function.

This general formulation applies to diverse networked systems, including power grid management, social network dynamics, and epidemic control. We focus on two representative and contrasting control tasks to demonstrate its utility.

\begin{example}[Epidemic Containment]
In this task, the goal is to minimize the spread of a disease by managing some control level (e.g., quarantine/vaccination intensity). The state of agent $i$ is $s_{i,t} = (h_{i,t}, c_{i,t})$, where $h_{i,t} \in \{0 \text{ (susceptible)}, 1 \text{ (infected)}\}$ is its health status and $c_{i,t} \in [0, 1]$ is its control value. The action $a_{i,t} \in \{-\Delta c, 0, \Delta c\}$ is to decrease, maintain, or increase the control value. The transition of the state $s_{i,t} \to s_{i,t+1}$ happens in two parts. First, the health status $h_{i,t}$ transitions probabilistically based on the \emph{current} state. Let $I_{i,t} = |\{j \in \mathcal{N}_i \setminus \{i\} : h_{j,t} = 1 \}|$ denote the number of infected neighbors. The probability of an agent being susceptible at the next timestep is given by: 
\begin{equation} P(h_{i,t+1} = 0 \mid s_{\mathcal{N}_i, t}) = 
    \begin{cases} \delta & \text{if } h_{i,t} = 1, \\ (1 - \beta(c_{i,t}))^{I_{i,t}} & \text{if } h_{i,t} = 0,
    \end{cases} 
    \label{epidemic_dynamics}
\end{equation} 
where $\delta$ is the fixed probability of recovery. The effective transmission rate $\beta(c_{i,t}) = (1 - \eta c_{i,t})\beta_0$ is a function of a base transmission rate $\beta_0$, the agent's current control state $c_{i,t}$, and the control effectiveness $\eta$ (e.g., for $\eta=0.9$, full-control provides at most 90\% reduction in infection probability for 1 infected neighbor).
Second, the control level updates deterministically based on the action: $c_{i,t+1} = \text{clip}(c_{i,t} + a_{i,t}, 0, 1)$. The reward function is structured to penalize the global infection rate and control costs.
\end{example}

\begin{example}[Rumor Spreading] 
In this task, the goal is to maximize the spread of information by managing advertising efforts. The state of agent $i$ is $s_{i,t} = (h_{i,t}, c_{i,t})$, where $h_{i,t} \in \{0 \text{ (unaware)}, 1 \text{ (aware)}\}$ is its awareness status and $c_{i,t} \in [0, 1]$ is its advertising level, which we call the ``boosting-factor.'' The action $a_{i,t} \in \{-\Delta c, 0, \Delta c\}$ is to decrease, maintain, or increase the boosting-factor. The awareness status transitions probabilistically, where the dynamics follow a ``viral marketing'' scenario with market saturation. Let $I_{i,t} = |\{j \in \mathcal{N}_i \setminus \{i\} : h_{j,t} = 1 \}|$ be the number of aware neighbors. The probability of an agent remaining unaware is then:
\begin{equation}
    P(h_{i,t+1} = 0 \mid s_t) =
    \begin{cases}
        0 & \text{if } h_{i,t} = 1, \\
        (1 - \beta(s_t, c_{i,t}))^{I_{i,t}} & \text{if } h_{i,t} = 0.
    \end{cases}
    \label{rumor_dynamics}
\end{equation}
The transmission probability $\beta(s_t, c_{i,t}) = c_{i,t} (1 - \bar{h}_t)^\kappa \beta_0$ now depends on the base spreading rate $\beta_0$, the boosting-factor state $c_{i,t}$, the fraction of aware nodes $\bar{h}_t = \frac{1}{N}\sum_{j=1}^{N} h_{j,t}$, and a saturation exponent $\kappa > 0$. Note that the saturation component in this example explicitly incorporates long-range interactions into the transition dynamics, which violates \Cref{eq:factorize}, but provides a contrasting variation to the purely local dynamics of the epidemic example. Finally, the boosting-factor updates deterministically: $c_{i,t+1} = \text{clip}(c_{i,t} + a_{i,t}, 0, 1)$. Here, a reward function is designed to encourage the aware state while penalizing the cost of the global boosting-factor usage.
\end{example}

These examples highlight the two core challenges we aim to address. These systems are vulnerable to \emph{long-range interactions}, as local agent states and actions have complex, non-local consequences. Further, they underscore the need for \emph{network generalizability}, as an ideal control policy should be transferable and effective both across local agent observations and across entire network topologies without requiring retraining.

Our proposed STACCA framework is designed to address these challenges by drawing from several methods in the literature. We now review the relevant background on MAPPO, the Transformer, and Graph Attention Networks (GATs).

\subsection{Background: Multi-Agent Proximal Policy Optimization}
MAPPO \cite{Yu2022SurprisingPPO} is an on-policy actor-critic algorithm that adopts the Centralized Training with Decentralized Execution (CTDE) paradigm \cite{kraemer2016multi}. Under CTDE, a centralized critic has access to global information (e.g., the joint state of the entire graph $\mathcal{G}$) during training, providing a more stable and informative learning signal. During execution, however, each agent $i$ employs a (shared) decentralized actor policy $\pi_{\theta}(a_i|o_i)$ to select actions using only its local observation $o_i$. Here, we have used parameter sharing, so all agents use the same policy neural network parameterized by $\theta$.

At each training round, suppose the current actor parameter is $\theta_{\text{old}}$. We collect a batch of trajectories $\mathcal{D}$ by executing this policy in the environment, where at each timestep $t$ in trajectory $\tau$, agent $i$ samples an action $a_{i,t}^\tau \sim \pi_{\theta_{\text{old}}}(\cdot|o_{i,t}^\tau)$ based on its local observation. The next actor parameter is obtained by optimizing the clipped surrogate objective over all agents, trajectories, and timesteps:
\begin{equation} \label{eq:clip_loss}
\begin{split}
L_{\pi}^{CLIP}(\theta) 
&= \hat{\mathbb{E}}_{i,\tau,t} \Big[
    \min \big(
        \rho_{i,t}^\tau(\theta)\, \hat{A}_{t}^\tau,\\
&\qquad\quad
        \text{clip}\!\left(\rho_{i,t}^\tau(\theta), 1 - \epsilon, 1 + \epsilon\right) \hat{A}_{t}^\tau
    \big)
\Big],
\end{split}
\end{equation}
where $\hat{\mathbb{E}}_{i,\tau,t}[\dots]$ denotes the empirical average over the samples $(s_{i,t}^\tau,a_{i,t}^\tau,o_{i,t}^\tau)$, $\rho_{i,t}^\tau(\theta)$ is the probability ratio $\frac{\pi_\theta(a_{i,t}^\tau|o_{i,t}^\tau)}{\pi_{\theta_{\text{old}}}(a_{i,t}^\tau|o_{i,t}^\tau)}$, $\epsilon$ is the clipping hyperparameter, and $\hat{A}_{t}^\tau$ is an estimate of the advantage function $A_{\pi_{\text{old}}}(s_{t}^\tau,a_{t}^\tau)$ (see \Cref{eq:advantage_function}) which is usually shared among agents at each timestep. This estimate is typically computed via Generalized Advantage Estimation (GAE) \cite{schulman2015high}, which uses the critic $V_\phi(s_{t}^\tau)$ as a baseline value function. The critic parameters $\phi$ are updated by minimizing the loss (e.g., mean squared error) between predicted values and GAE returns $R_{t}^\tau  = \hat{A}_t^\tau + V_{\phi_{\text{old}}}(s_t^\tau)$ (see \cite{schulman2015high}): 
\begin{equation} L_{V}(\phi) = \hat{\mathbb{E}}_{\tau,t} \left[ \left( V_{\phi}(s_{t}^\tau) - R_{t}^\tau \right)^2 \right]. 
\end{equation}
This process of collecting trajectories and updating the actor and critic is repeated for multiple training iterations.

\subsection{Background: Transformers and Self-Attention}
\label{Transformer}

The Transformer architecture has been shown to be especially powerful in capturing long-range dependencies through the self-attention mechanism \cite{Vaswani2017AttentionAllYouNeed}. Given a set of input embeddings $X = [x_1, \dots, x_N]^\top \in \mathbb{R}^{N \times d_{\text{in}}}$, they are first projected into queries ($Q \in \mathbb{R}^{N \times d_k}$), keys ($K \in \mathbb{R}^{N \times d_k}$), and values ($V \in \mathbb{R}^{N \times d_v}$) using learned weight matrices $W_Q, W_K \in \mathbb{R}^{d_{\text{in}} \times d_k}$ and $W_V \in \mathbb{R}^{d_{\text{in}} \times d_v}$:
\begin{equation} \label{eq:att_mat}
Q = XW_Q, \quad K = XW_K, \quad V = XW_V.
\end{equation}
The core of the mechanism is computing pairwise scores between each query and all keys, which are then scaled and normalized via softmax to produce attention weights. The final output is a weighted sum of the values:
\begin{equation}
\text{Attention}(Q, K, V) = \underbrace{\text{softmax}\left(\frac{QK^\top}{\sqrt{d_k}}\right)}_{\text{Attention Weights}}V
\end{equation}
where $d_k$ denotes the dimensionality of the key (and query) vectors. This mechanism's global receptive field is critical for a centralized MARL critic to model system-wide, long-range interactions. Multi-head attention performs this computation in parallel, capturing diverse interaction patterns that are especially useful for a shared actor learning a generalizable policy. Furthermore, recent theoretical work investigating self-attention through the lens of interacting entities has provided a strong motivation for its application in modeling the complex dynamics of multi-agent systems \cite{Ustaomeroglu2025HyperSelfAttention}.

\subsection{Background: Graph Attention Networks}
\label{GAT}
While the global self-attention mechanism is powerful, most networked systems exhibit a known graph structure. To leverage this topology as an inductive bias, we employ GATs, which adapt self-attention to graph domains \cite{velickovic2018graph}.

Similar to the Transformer, given a set of input node embeddings $X = [x_1, \dots, x_N]^\top$, we first project them using a learned weight matrix $W \in \mathbb{R}^{d_k \times d_{\text{in}}}$. In the standard GAT formulation, this single projection serves as both the query and key transformations (i.e., $q_i = Wx_i$ and $k_j = Wx_j$). However, instead of attending to all other nodes in the system, each agent $i$ computes attention coefficients $\alpha_{ij}$ exclusively over its connected neighbors $j \in \mathcal{N}_i$.  

The unnormalized attention score $e_{ij}$ between a query $q_i$ and key $k_j$ is typically computed using an additive mechanism parameterized by a weight vector $\mathbf{a} \in \mathbb{R}^{2d_k}$ and a LeakyReLU activation:
\begin{equation}
e_{ij} = \text{LeakyReLU}\left(\mathbf{a}^\top [q_i  \|  k_j]\right)
\end{equation}
where $\|$ denotes concatenation. These scores are then normalized via the softmax function over all neighbors:
\begin{equation}
\alpha_{ij} = \frac{\exp(e_{ij})}{\sum_{k \in \mathcal{N}_i} \exp(e_{ik})}
\end{equation}

The final output for node $i$ is the weighted sum of the projected features of its neighbors, often followed by a nonlinear activation function $\sigma$:
\begin{equation}
x_i' = \sigma\left(\sum_{j \in \mathcal{N}_i} \alpha_{ij} W x_j\right)
\end{equation}

This approach combines attention-based weighting with the strong structural prior of the graph. For our shared actor, this allows agents to learn a generalizable, structure-aware policy by dynamically weighting information from relevant neighbors. For the centralized critic, stacking GAT layers enables information to propagate across the graph, allowing it to build a global value estimate that respects the underlying network topology (See \Cref{methods:arch}).

\section{Conditions for Long-Range Interactions}
\label{sec:theory}
In this section, we address Question 1 posed in \Cref{intro}: \emph{Can we better understand long-range interactions in networked systems?} While previous work demonstrated that local interactions exhibit an exponential decay property under certain ergodicity assumptions \cite{qu2020scalable}, preventing their effects from propagating to distant nodes, we study the inverse regime. In \Cref{sec:theory_mean_field}, we take a stability perspective and analyze the conditions under which perturbations amplify and persist in the mean-field approximation of the system.

In \Cref{sec:empirical}, we connect our stability analysis of the mean-field system's Jacobian matrix $J$ to long-range interactions in the true system. We present an empirical study showing that (1) the spectral radius of $J$, $\rho(J)$, is strongly correlated with a system's propensity for long-range interactions, and (2) perturbations more aligned with the unstable eigenspace of $J$ produce stronger long-range interactions. 

Our results, along with prior work analyzing the limitations of message-passing neural networks (MPNNs) for modeling long-range interactions, motivate a transformer-based MARL solution, which we discuss in \Cref{sec:transformer}.

\subsection{Mean-Field Analysis for Long-Range Interactions}
\label{sec:theory_mean_field}
\subsubsection{Setup}
 We consider the networked MDP framework discussed in \Cref{problem_formulation}. By assuming a fixed, stationary joint policy $\boldsymbol{\pi}$, the system reduces to a Markov Chain where the state of each node evolves according to a local transition kernel conditioned solely on the state of its closed neighborhood $\mathcal{N}_i$: $P_i^{\boldsymbol{\pi}}(s_{i,t+1} \mid s_{\mathcal{N}_i,t})$.\footnote{For brevity, we omit the $\boldsymbol{\pi}$ superscript in subsequent notation.} We study long-range interactions by analyzing this Markov Chain from a stability perspective. In particular, we analyze the onset of instability starting from a normal equilibrium state. This state is analogous to a disease-free equilibrium (DFE) in an epidemic setting, and it is defined explicitly below.

\begin{definition}[Normal Equilibrium]
\label{def:equilibrium}
A configuration $s^* = (s_1^*, \dots, s_N^*)$ is a normal equilibrium if it is an absorbing state of the local dynamics: $P_i(s_i^* \mid s_{\mathcal{N}_i}^*) = 1$ for all $i \in \mathcal{N}$. The corresponding marginal distribution for each node is an indicator function, $p_i^*(x_i) = \mathbf{1}_{\{x_i = s_i^*\}}$, meaning the unperturbed joint distribution is the product of these independent marginals.
\end{definition}

\subsubsection{Mean-field Factorization} Let $p_{i,t}(x_i)$ denote the marginal distribution of node $i$ at timestep $t$, and let $p_{\mathcal{N}_i,t}(x_{\mathcal{N}_i}) := \Pr(s_{\mathcal{N}_i,t} = x_{\mathcal{N}_i})$ denote the joint distribution of the state of the nodes in $\mathcal{N}_i$. Then, the exact marginal recursion is given by:
\begin{equation} \label{eq:exact_marginal_recursion}
p_{i,t+1}(x_i') = \sum_{x_{\mathcal{N}_i}}
P_i(x_i' \mid x_{\mathcal{N}_i}) \, p_{\mathcal{N}_i,t}(x_{\mathcal{N}_i}).
\end{equation}
Tracking the exact evolution of $p_{i,t}$ is computationally intractable because the joint distribution over the neighborhood, $p_{\mathcal{N}_i,t}$, requires the joint distribution of the 2-hop neighborhood at $t-1$, the 3-hop neighborhood at $t-2$, and so forth. This rapid expansion eventually encompasses the entire network, requiring us to maintain the full global joint distribution. To achieve a tractable formulation, we use a mean-field approximation.

\begin{assumption}[Mean-Field Factorization] \label{assump:mean_field}
At any time step $t$, the joint distribution of the neighborhood is approximated by the product of independent marginals:
\begin{equation} \label{eq:mf_assumption}
p_{\mathcal{N}_i,t}(x_{\mathcal{N}_i}) \approx \prod_{j \in \mathcal{N}_i} p_{j,t}(x_j).
\end{equation}
\end{assumption}

Under Assumption \ref{assump:mean_field}, the marginal recursion simplifies to a discrete-time nonlinear dynamical system governing the individual marginal probabilities:
\begin{equation} \label{eq:mf_recursion}
p_{i,t+1}^{\mathrm{MF}}(x_i') = \sum_{x_{\mathcal{N}_i}} P_i(x_i' \mid x_{\mathcal{N}_i}) \prod_{j \in \mathcal{N}_i} p_{j,t}^{\mathrm{MF}}(x_j), \quad \forall i \in \mathcal{N}.
\end{equation}

We denote the global mean-field evolution as $p_{t+1}^{\mathrm{MF}} = F(p_t^{\mathrm{MF}})$, where $F$ is the nonlinear dynamics operator. To assess the stability of the normal equilibrium, we linearize $F$ about $p^*$.

Because each marginal lies on the probability simplex, $\sum_{x_k \in \mathcal{S}_k} p_{k,t}(x_k) = 1$, its coordinates are not independent: fixing the non-reference probabilities determines the reference one via $p_{k,t}(s_k^*) = 1 - \sum_{x_k \neq s_k^*} p_{k,t}(x_k)$. We therefore designate the normal state $s_k^*$ as the reference for each node $k$ and linearize in the remaining (free) coordinates.

\begin{definition}[Constrained Jacobian]
\label{def:constrained_jacobian}
Let $J$ be the Jacobian of $F$ in the free (non-reference) coordinates, evaluated at $p^*$, with block entries $J_{ik}(x_i', x_k)$ defined for $x_i' \neq s_i^*$ and $x_k \neq s_k^*$. Eliminating the dependent reference coordinate through the simplex constraint yields, by the chain rule,
\begin{equation}
\label{eq:constrained_general}
J_{ik}(x_i', x_k)
= \left.\frac{\partial p_{i,t+1}^{\mathrm{MF}}(x_i')}{\partial p_{k,t}^{\mathrm{MF}}(x_k)}\right|_{p^*}
- \left.\frac{\partial p_{i,t+1}^{\mathrm{MF}}(x_i')}{\partial p_{k,t}^{\mathrm{MF}}(s_k^*)}\right|_{p^*},
\end{equation}
where the second term is the contribution of the reference state induced by the constraint.
\end{definition}

\begin{remark}[Reduction at the Normal Equilibrium]
\label{rem:jacobian_reduction}
At the normal equilibrium $p^*$, the reference-state term in \eqref{eq:constrained_general} vanishes, and the constrained Jacobian reduces to the single-perturbation response
\begin{equation}
\label{eq:normal_jacobian}
J_{ik}(x_i', x_k)=
\begin{cases}
P_i\!\left(x_i' \mid s_{\mathcal{N}_i \setminus \{k\}}^*,\, x_k\right)
& \text{if } k \in \mathcal{N}_i,\\[2pt]
0 & \text{otherwise.}
\end{cases}
\end{equation}
\end{remark}

This follows because $s^*$ is an absorbing state (\Cref{def:equilibrium}), $P_i(s_i^* \mid s_{\mathcal{N}_i}^*) = 1$, and hence $P_i(x_i' \mid s_{\mathcal{N}_i}^*) = 0$ for every $x_i' \neq s_i^*$. The subtracted term in \eqref{eq:constrained_general} therefore vanishes, leaving \eqref{eq:normal_jacobian}.

\begin{proposition}\label{prop:long_range} For the mean-field system with linearized dynamics, $J$, the equilibrium $p^*$ is locally exponentially stable if $\rho(J)<1$ and locally unstable if $\rho(J)>1$.
\end{proposition}

The above follows from Lyapunov's Indirect Method. An unstable normal equilibrium under the mean-field dynamics gives an approximate condition for whether a perturbation is likely to persist over time under the true dynamics, but it does not guarantee \emph{spatial} long-range interactions. To further study this condition under the true system dynamics, we perform an empirical analysis in \Cref{sec:empirical}. To preface this analysis, we derive the linearized mean-field dynamics for the epidemic example below.

\subsubsection{Epidemic Example}
\label{theory_mean_field_example}
To illustrate \Cref{prop:long_range}, we apply it to a simplified version of the Epidemic Containment task introduced in \Cref{problem_formulation}. Let the probability of node $i$ being infected at time $t$ be $p_{i,t} = \Pr(h_{i,t} = 1)$. We assume a fixed control policy $\pi_i$, yielding a constant effective transmission rate $\beta_i = \beta(\pi_i)$. The normal equilibrium is the disease-free state, where $h_i^* = 0$ (susceptible) for all $i$, yielding $p_i^* = 0$. Under the mean-field approximation, the probability that node $i$ avoids infection from its neighbors is $\prod_{j \in \mathcal{N}_i \setminus \{i\}} (1 - \beta_i p_{j,t}^{\mathrm{MF}})$. The marginal dynamics are therefore:
\begin{equation}p_{i,t+1}^{\mathrm{MF}} = p_{i,t}^{\mathrm{MF}} (1 - \delta) + (1 - p_{i,t}^{\mathrm{MF}}) \left[ 1 - \prod_{j \in \mathcal{N}_i \setminus \{i\}} (1 - \beta_i p_{j,t}^{\mathrm{MF}}) \right].
\end{equation}

We compute the Jacobian $J$ evaluated at the normal equilibrium $p^* = \mathbf{0}$.
For the diagonal entries ($k = i$):
\begin{equation}
\frac{\partial p_{i,t+1}^{\mathrm{MF}}}{\partial p_{i,t}^{\mathrm{MF}}}\Big|_{p^*=\mathbf{0}} = (1 - \delta) - \left[ 1 - \prod_{j} (1 - \beta_i \cdot 0) \right] = 1 - \delta.
\end{equation}
For the off-diagonal entries corresponding to neighbors ($k \in \mathcal{N}_i \setminus \{i\}$):
\begin{equation}
\begin{aligned}
\frac{\partial p_{i,t+1}^{\mathrm{MF}}}{\partial p_{k,t}^{\mathrm{MF}}}\Bigg|_{p^*=\mathbf{0}}
&= (1-0)\left[
- \frac{\partial}{\partial p_{k,t}^{\mathrm{MF}}}
\prod_{j \in \mathcal{N}_i \setminus \{i\}}
(1-\beta_i p_{j,t}^{\mathrm{MF}})
\right]\Bigg|_{p^*=\mathbf{0}} \\
&= \left[ \beta_i \prod_{j \in \mathcal{N}_i \setminus \{i,k\}} (1-\beta_i p_{j,t}^{\mathrm{MF}}) \right]\Bigg|_{p^*=\mathbf{0}} = \beta_i .
\end{aligned}
\end{equation}
Thus, the Jacobian matrix governing the linear propagation of infections is:
\begin{equation} \label{eq:epidemic_jacobian}
J = (1 - \delta)I + B \mathbf{A},
\end{equation}
where $I$ is the identity matrix, $\mathbf{A}$ is the adjacency matrix, and $B = \operatorname{diag}(\beta_1, \dots, \beta_N)$ reflects the effective transmission rates for each node under their control value. Observe that $J$ encodes both \emph{structural} information through $\mathbf{A}$ as well as \emph{dynamics} information through $\delta$ and $B$. Interestingly, when the local control intensities are uniform ($\beta_1 = \beta_2 = \dots = \beta_N = \beta$), the eigenvalues of $J$ are equal to $(1 - \delta) + \beta \lambda_k(\mathbf{A})$, where $\lambda_k(\mathbf{A})$ are the eigenvalues of $\mathbf{A}$, and the eigenvectors are the same as those of $\mathbf{A}$.

\subsection{Empirical Analysis for Long-Range Interactions}
\label{sec:empirical}

While \Cref{prop:long_range} addresses the local stability of the normal equilibrium under mean-field dynamics, analyzing the system outside of the normal equilibrium by tracking the evolution of the Jacobian through the nonlinear regime is analytically intractable. Furthermore, while the mean-field system provides an approximation of the true system, we have yet to explore how it translates to the true system. To bridge these gaps, we conduct an empirical analysis.

We use a simplified version of the epidemic environment introduced in \Cref{problem_formulation} that excludes control actions and uses a fixed transmission rate ($\beta_i = \beta$ for all $i$) for all nodes. We run simulations over randomly generated network topologies with varying spectral radii for the linearized mean-field dynamics, $\rho(J)$. To ensure our networks capture a diverse range of values for $\rho(J)$, we generate 500 distinct Erdős–Rényi networks \cite{erdos1959random} ($N=500$ nodes) while varying the parameter for edge creation probability, $p$. The dynamics are governed by a transmission rate $\beta = 0.1$ and recovery rate $\delta = 0.5$.

We compare perturbations that strongly align with the unstable eigenspace of the linearized mean-field dynamics, $J$, to those that weakly align. As an approximate measurement of this alignment, we use eigenvector centrality \cite{bonacich1972factoring}. As shown in the derivation in \Cref{theory_mean_field_example}, the eigenvectors of $J$ are equivalent to the eigenvectors of the adjacency matrix $\mathbf{A}$. Thus, we compare two initial seeding strategies. \textbf{(1) Low Eigenvector Centrality Seeding:} 1 permanently infected node is randomly sampled from the 10 nodes with the lowest magnitude components in the dominant eigenvector of $\mathbf{A}$ (and $J$). \textbf{(2) High Eigenvector Centrality Seeding:} 1 permanently infected node is randomly sampled from the 10 nodes with the highest magnitude components in the dominant eigenvector of $\mathbf{A}$ (and $J$).

This seeding strategy is a practical proxy for alignment (or misalignment) with the unstable eigenspace of $J$. While the high centrality seeds are strongly aligned with the dominant unstable direction, the low centrality seeds are only misaligned with the dominant eigenvector, but they may still possess non-trivial projections onto other unstable eigenvectors (i.e., where $\lambda > 1$) in highly unstable networks. Furthermore, we emphasize that we use a permanently infected seed for our simulated perturbations. While this acts as a continuous forcing rather than a pure initial condition, it prevents trivial 1-step recoveries and ensures a sustained perturbation, allowing us to clearly observe whether the network amplifies or suppresses the cascade from a specified structural origin.

\begin{figure}[!ht]
    \centering
    \begin{subfigure}[c]{0.49\linewidth}
        \centering
        \includegraphics[width=\linewidth]{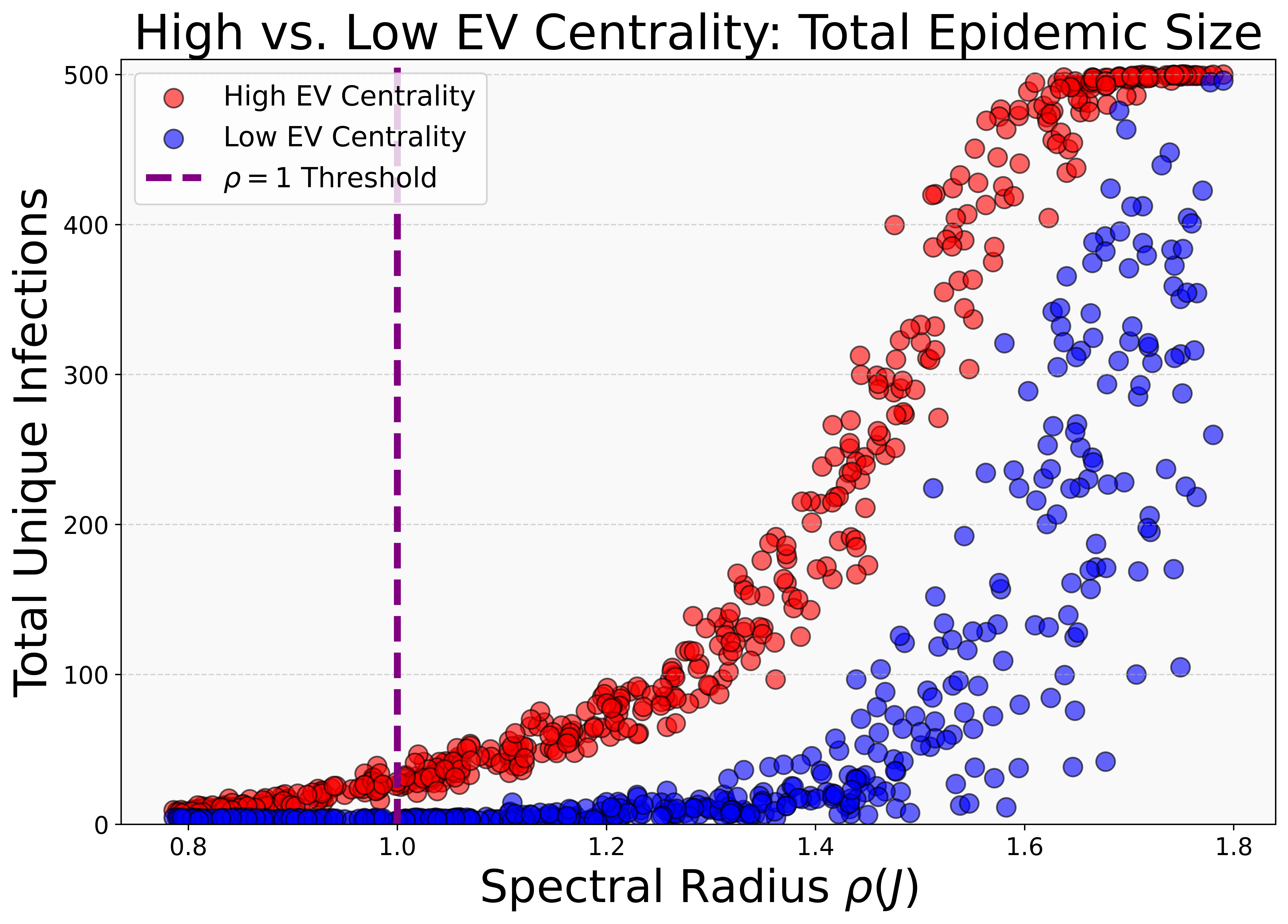}
    \end{subfigure}
    \begin{subfigure}[c]{0.49\linewidth}
        \centering
        \includegraphics[width=\linewidth]{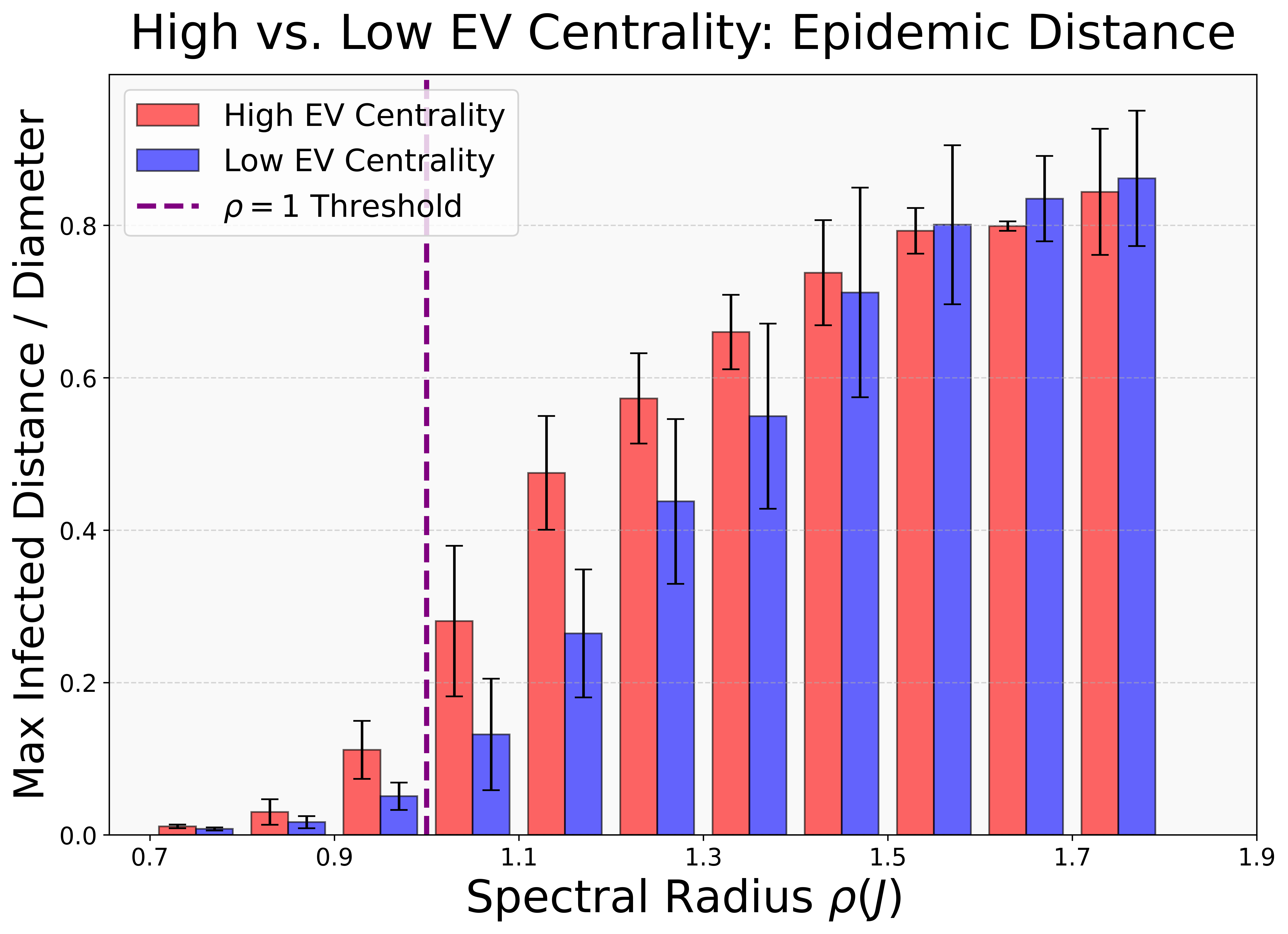}
    \end{subfigure}
    \caption{Comparison of low eigenvector centrality (EV) seeding to high EV seeding for epidemic spreading. We plot the total infected nodes (left) and the distance traveled from the initial seed (right) over 50 timesteps.}
    \label{fig:epidemic_sim}
\end{figure}

For each topology, we run simulations for 50 timesteps and record (1) the total number of infected nodes over the entire horizon and (2) the maximum distance the infection travels from the initial seed, averaged across 5 independent runs (See \Cref{fig:epidemic_sim}). Our results highlight two key takeaways:

\textbf{1. $\rho(J)$ is strongly correlated with long-range interactions.} $\rho(J)$ is strongly correlated with both the total volume of infected nodes and the relative epidemic distance from the initial seed. This demonstrates that the stability condition under the mean-field dynamics also serves as a powerful predictor for the severity of long-range interactions under the true system dynamics.

\textbf{2. The origin of the perturbation has a significant impact on spatial long-range interactions.} The results reveal a stark disparity in long-range interactions depending on the origin of the perturbation. Infections originating at mathematically ``safe'' nodes largely fail to achieve significant spread, particularly when compared to the less ``safe'', high eigenvector centrality seeds. In highly unstable networks where $\rho(J) \gg 1$, low eigenvector centrality seeds do occasionally trigger larger outbursts, as these nodes may still align with other non-dominant unstable eigenvectors. Conversely, high eigenvector centrality seeds consistently exploit the network's topological instability to drive spatial long-range effects.

This motivates a network control approach that considers the nodes' varying levels of vulnerability to long-range interactions. If a perturbation occurs at a node that is highly aligned with the unstable eigenspace of $J$, distant agents must be able to recognize this critical state and preemptively adjust their local policies before the cascade materializes. This necessitates a communication mechanism capable of delivering high-fidelity signals across the network, directly motivating our transformer-based architecture discussed in \Cref{sec:transformer}.

\subsection{Motivation for Transformer-based Solution}
\label{sec:transformer}

\Cref{sec:theory_mean_field} and \Cref{sec:empirical} have shown that nodes can have varying levels of vulnerability to long-range interactions. This poses a significant communication challenge for decentralized network control. In order to suppress a long-range cascade before it fully materializes, nodes must be able to distinguish between vulnerability levels and recognize when highly vulnerable nodes are perturbed (e.g., infected) so that they adjust their local control policies accordingly. This requires a communication architecture capable of preserving high-fidelity signals over long network distances.

Standard Graph Neural Networks (GNNs), such as GATs (see \Cref{GAT}), rely on the message-passing paradigm, where information is propagated strictly along the edges of the underlying topology. While this is an excellent method for capturing local interactions and the structural inductive bias of the network topology, message-passing mechanisms are fundamentally ill-equipped to facilitate the long-range signaling required in our setting. Recent theoretical work has demonstrated that as the receptive field of a GNN grows to capture distant nodes, the network suffers from ``over-squashing" \cite{topping2021understanding, alon2021bottleneck}.

\begin{figure}[!ht]
    \centering
    \includegraphics[width=0.9\linewidth]{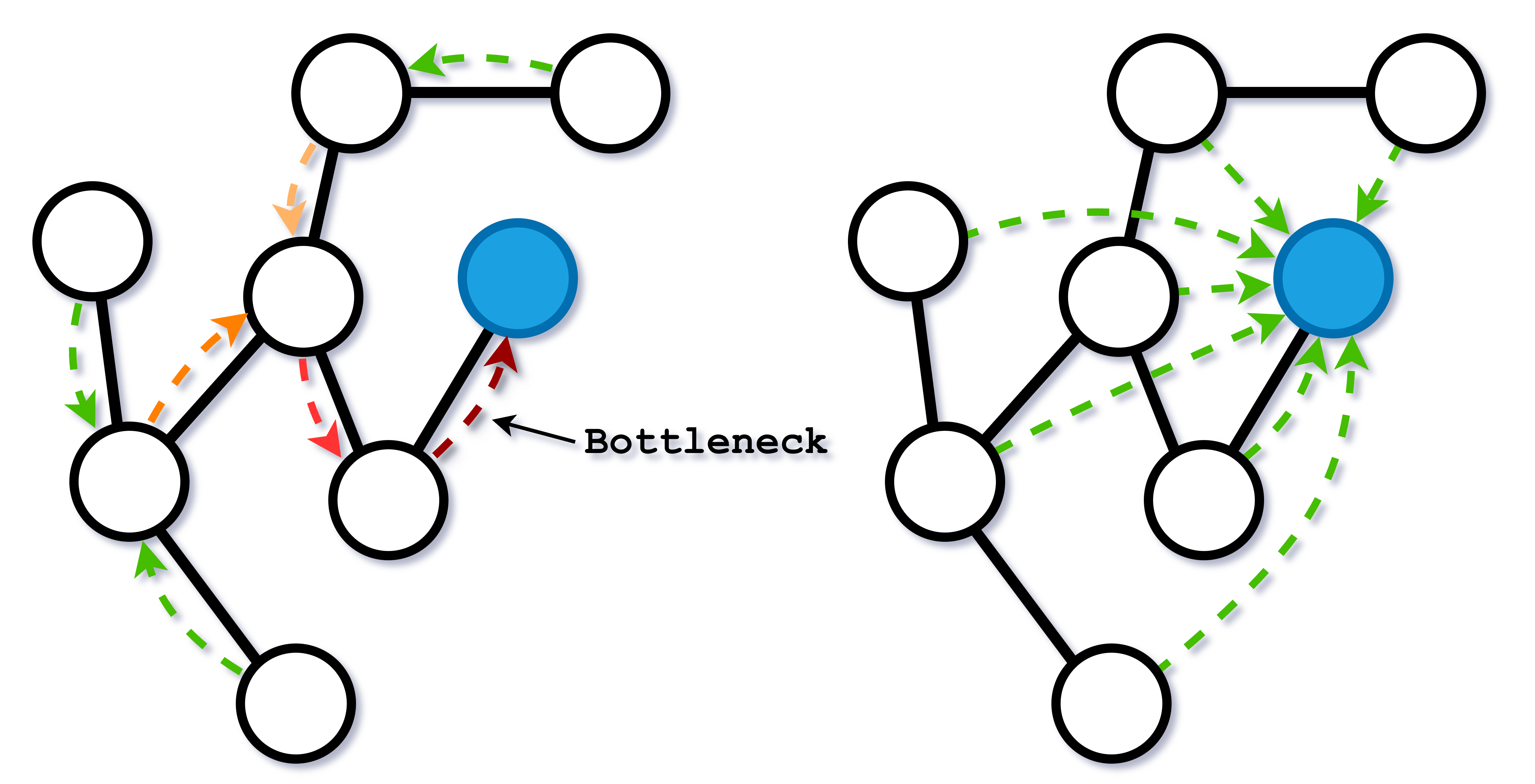}
    \caption{MPNN (left) vs. full self-attention (right) information routing architectures. In MPNNs, signals from distant nodes are compressed and distorted (indicated by arrow colors) as they are forced through structural graph bottlenecks. The self-attention mechanism constructs a fully connected communication graph, allowing for direct, high-fidelity signals between any two nodes regardless of network distance.}
    \label{fig:oversquash}
\end{figure}

In the context of our epidemic containment task, over-squashing implies that if a node becomes infected, its warning signal will be heavily distorted or entirely suppressed by the time it reaches distant susceptible nodes through a standard MPNN. This loss of signal fidelity is visualized in the left panel of \Cref{fig:oversquash}, and it can lead to a failure in proactive global control when highly vulnerable nodes are infected.

This directly motivates the use of a full self-attention mechanism, such as the Transformer \cite{Vaswani2017AttentionAllYouNeed} (see \Cref{Transformer}). As shown in the right panel of \Cref{fig:oversquash}, by computing pairwise attention scores between all nodes regardless of network distance, the Transformer effectively constructs a fully connected communication graph. This allows any node in the network to directly attend to the state of all other nodes without routing through topological bottlenecks. Consequently, a Transformer-based architecture prevents the over-squashing of critical warning signals, enabling the agents to exploit global structural awareness and effectively suppress long-range cascading failures.

However, while full self-attention effectively eliminates over-squashing, it inherently lacks the structural inductive bias on the underlying network topology. To achieve the best of both worlds, we propose STACCA, which combines an MPNN (GAT architecture) with the Transformer architecture. As detailed in our methodology (\Cref{methods}) and validated by our experimental results, this hybrid approach exploits the network structure while simultaneously maintaining high-fidelity global communication channels.

\section{Methods}
\label{methods}

We now turn to Question 2 posed in \Cref{intro}: \emph{Can we create a MARL solution for networked systems that handles long-range interactions and is network generalizable?} To answer this question, we propose STACCA, a Transformer-based MARL framework. In \Cref{methods:arch}, we introduce (1) a Graph Transformer Critic architecture designed to learn global, long-range dependencies among agents to guide local policy learning and (2) a Graph Transformer Actor architecture designed to adapt to heterogeneous local neighborhoods for network generalizability. Furthermore, in addressing the above challenges, we require training on relatively large graphs to both capture long-range interactions and expose the actor to diverse local network topologies. This, in turn, exacerbates the significant challenge of \emph{multi-agent credit assignment}. To address this, we introduce a novel counterfactual advantage, which we cover in \Cref{methods:counterfactual}. Finally, we present the complete STACCA framework in \Cref{methods:method}.

\subsection{Graph Transformer Actor and Critic Architectures}
\label{methods:arch}

\begin{figure}[!ht]
    \centering
    \begin{minipage}[!ht]{0.4\textwidth}
        \centering
        \includegraphics[width=\textwidth]{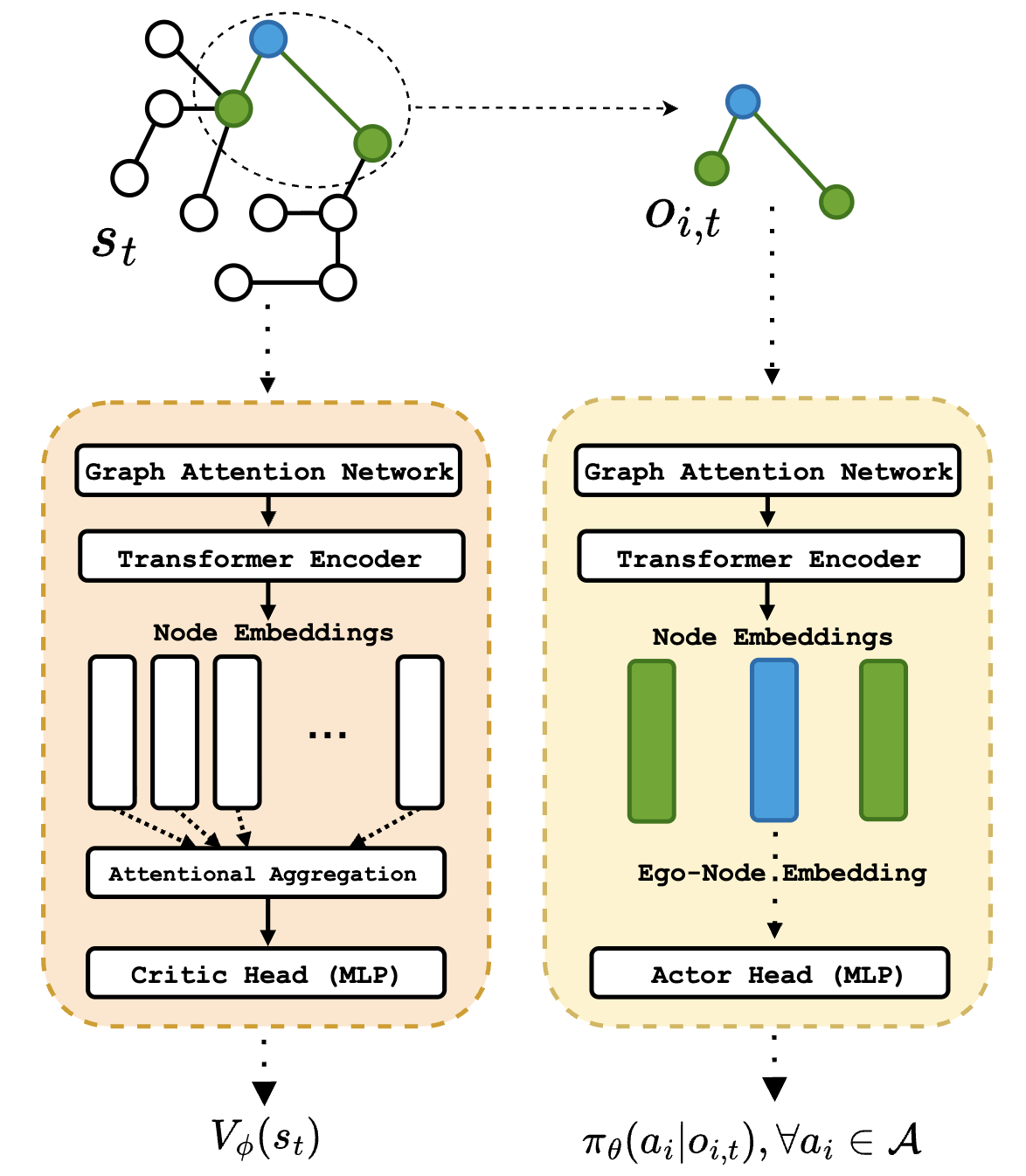}
        \caption{Graph Transformer Critic (left) and Actor (right) Architectures. Node $i$ is represented by the blue node. Node $i$'s 1-hop neighborhood (green nodes) is its local observation at time $t$, $o_{i,t}$.}
        \label{fig:arch}
    \end{minipage}
\end{figure}

To address the challenges of modeling long-range interactions and learning a network-generalizable shared policy, we introduce critic and actor networks, respectively, based on a hybrid Graph Transformer architecture. Both networks combine the strengths of GATs for leveraging the explicit graph structure with the power of Transformer self-attention for learning (potentially) long-range global dependencies. However, while sharing this foundational structure, the actor and critic differ in their scope and final aggregation mechanisms, tailored to their distinct roles (see \Cref{fig:arch}).

The \emph{centralized Graph Transformer Critic} processes the entire graph to learn the global state-value function $V_{\phi}(s)$. Each node \(i\) begins with raw features \(s_i\), which are first projected through a multi-layer perceptron (MLP) embedding transformation \(f(\cdot)\) to obtain initial node embeddings \(e_i = f(s_i)\). These embeddings are then refined by a stack of GAT layers, producing structure-aware representations \(h_i = g(e_i)\). The resulting collection of embeddings \(H = [h_1, \dots, h_N]^{\top}\) serves as the Transformer Encoder input (\(X\) in \Cref{eq:att_mat}). This two-stage process of extracting a structural inductive bias through the GAT layers, then refining the embeddings with a Transformer Encoder to model long-range interactions serves as the backbone for STACCA's critic and actor architectures. Following the GAT and Transformer layers, it employs an attentional aggregation mechanism to produce a single embedding vector representing the entire graph state. This learned, weighted average of all node embeddings creates a holistic representation, which is then passed to an MLP (critic head) to produce the final scalar value estimate.

The \emph{decentralized Graph Transformer Actor} is a shared policy network \(\pi_{\theta}(a_i|o_i)\) applied independently at each node to produce an action distribution. The overall structure is similar to that of the critic, except the input is agent \(i\)’s local observation \(o_i\). The attention-based architecture allows the shared policy to adapt naturally to heterogeneous local topologies. After neighborhood processing, the actor selects the ego-node embedding (embedding corresponding to node $i$), which is now infused with topological context from the GAT and Transformer layers, as input to an MLP (policy head), enabling specialized, context-sensitive behaviors. In contrast, an averaging mechanism, while increasing stability in the critic, can potentially push agents into more uniform behaviors in the actor by "smoothing out" the local context.

Regarding computational complexity, the global self-attention mechanism in the centralized critic scales quadratically, $O(N^2)$, with respect to the number of agents $N$. While this could be a bottleneck for massive training networks, modern GPU parallelization renders the computational overhead practically negligible for the networks used in this work ($N = 50$ during training). Furthermore, because the critic is centralized and only used during training, this $O(N^2)$ cost does not impact decentralized execution, even when the evaluation network is scaled to an arbitrarily large $N$. 

\subsection{Counterfactual Advantage Calculation}
\label{methods:counterfactual}

Counterfactual reasoning \cite{Foerster2018COMA} is a powerful technique used to address the multi-agent credit assignment problem by isolating each agent's contribution to the shared reward. It does so through an advantage formulation that compares an agent's action to an individualized counterfactual baseline---what would have happened if the agent had acted differently---as opposed to the global value function baseline shown in \Cref{eq:advantage_function}. However, existing methods that use this idea typically rely on learning a joint action-value critic $Q_\psi(s,a)$ \cite{Foerster2018COMA}. These methods scale poorly with the number of agents and are incompatible with the more stable state-value critic $V_\phi(s)$ used in MAPPO.

\begin{figure}[!ht]
    \centering
    \begin{minipage}[!ht]{0.45\textwidth}
        \centering
        \includegraphics[width=\textwidth]{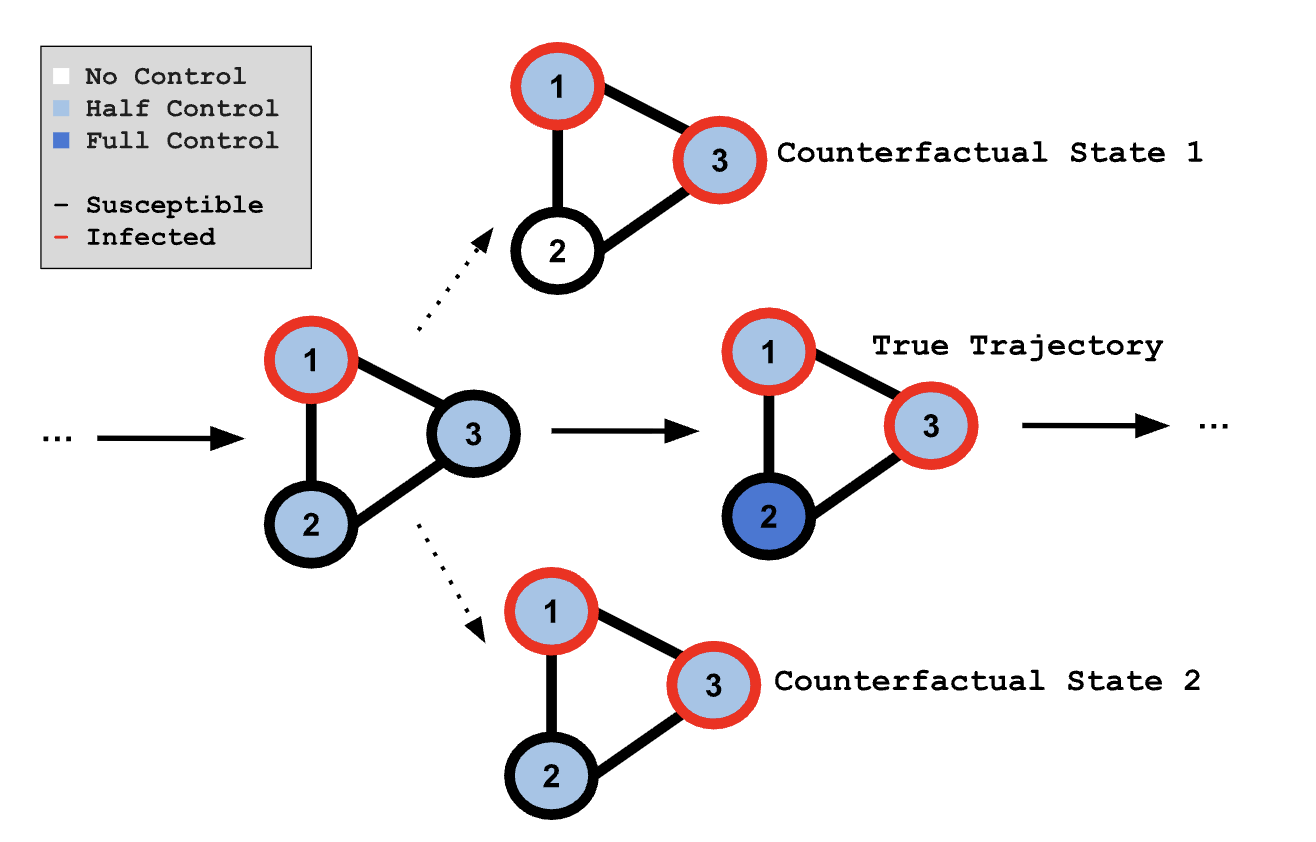}
        \caption{One-step trajectory branching for Node 2 in a 3-node epidemic containment environment. }
        \label{fig:counterfactual}
    \end{minipage}
\end{figure}

We introduce a novel calculation procedure and baseline for counterfactual advantages that are compatible with MAPPO's state-value critic and are effective even with a large number of agents. While a state-value critic $V_\phi(s)$ enhances stability compared to a joint action-value critic $Q_\phi(s,a)$, it does not directly provide the action-dependent values needed for counterfactual reasoning. Our method extracts this information using an efficient, three-step procedure:

\textbf{1. One-step Trajectory Branching.} For each agent \(i\) at each timestep \(t\), we generate one-step counterfactuals. We iterate through every possible action \(a'_i \in \mathcal{A}_i\) and compute the counterfactual next state \(s'_{t+1}(a'_i)\) that would have occurred if agent \(i\) had taken action \(a'_i\) while all other agents' actions \(a_{-i,t}\) remained the same (see \Cref{fig:counterfactual}). For continuous actions, this can still be accomplished by discretizing or sampling from the action space.

\textbf{2. Compute Advantages.}  We use the centralized critic \(V_\phi\) to compute the counterfactual advantage. This advantage is the difference between the one-step return of the agent's taken action and the policy-weighted average of the one-step returns of all its possible actions: 
\begin{equation} \label{eq:cf_adv}
\begin{split}
    \hat{A}^{CF}_{i,t} ={}& [r_t + \gamma V_\phi(s_{t+1})] \\
    & - \sum_{a'_i \in \mathcal{A}_i} \pi_\theta(a'_i|o_{i,t}) [r'_{t}(a'_i) + \gamma V_\phi(s'_{t+1}(a'_i))],
\end{split}
\end{equation}
where \(r_t\) and \(r'_{t}(a'_i)\) are the rewards from the actual and counterfactual actions, respectively.\footnote{\Cref{eq:cf_adv} shows the counterfactual advantage formulation with one-step returns for the sake of clarity. In the implementation, GAE returns were used (see \cite{schulman2015high}).} This provides each agent with a personalized baseline, isolating its individual contribution.

\textbf{3. Amplify Credit-Assignment Signal.} A stable, low-variance critic may produce only small differences in value between these counterfactual branches. To amplify these crucial learning signals, we introduce a \emph{timestep-level normalization} step. At each timestep, we normalize the computed advantages across all \(N\) agents:
\begin{equation} \label{eq:cf_norm}
\tilde{A}^{CF}_{i,t} = \frac{\hat{A}^{CF}_{i,t} - \mu_{A_t}}{\sigma_{A_t} + \epsilon},
\end{equation}
where \(\mu_{A_t}\) and \(\sigma_{A_t}\) are the mean and standard deviation of the advantages of all agents at timestep \(t\). This procedure re-scales each agent's contribution relative to the team's average performance at timestep \(t\) and is particularly effective in systems with many agents, where these cross-agent statistics are meaningful. 

Finally, to integrate these agent-specific normalized advantages into the clipped surrogate objective \eqref{eq:clip_loss}, we use the following updated objective:
\begin{equation} \label{eq:clip_loss_cf}
\begin{split}
L_{\pi}^{CLIP,CF}(\theta) 
&= \hat{\mathbb{E}}_{i,\tau,t} \Big[
    \min \big(
        \rho_{i,t}^\tau(\theta)\, \tilde{A}^{\tau,CF}_{i,t},\\
&\qquad\quad
        \text{clip}\!\left(\rho_{i,t}^\tau(\theta), 1 - \epsilon, 1 + \epsilon\right) \tilde{A}^{\tau,CF}_{i,t}
    \big)
\Big].
\end{split}
\end{equation}

This procedure is computationally efficient. In our networked environments, an agent's action has a localized effect, so constructing each counterfactual state is an \(O(1)\) operation. The total complexity per timestep is thus \(O(AN)\) for all agents (where \(A=|\mathcal{A}_i|\)), which does not increase the asymptotic complexity of the MAPPO training loop.\footnote{In our experiments, we add a small, linearly increasing bias to the advantages (starting at 0 and increasing by 0.1 every 10 episodes) that occasionally helped stabilize training in later stages.}

Our approach builds on the principles of counterfactual reasoning, most notably introduced by COMA \cite{Foerster2018COMA}. The one-step trajectory branching with timestep-level normalization efficiently extracts action-dependent values from a state-value critic $V_\phi(s)$ and processes them into effective learning signals.  This enables compatibility with a broader variety of MARL algorithms and scales well to many agents.

\subsection{Proposed Method: STACCA}
\label{methods:method}
We now present the complete STACCA framework integrated into the base MAPPO training loop; however, STACCA is designed as a modular framework, and its core components can be integrated into various actor-critic MARL algorithms. Specifically, the Graph Transformer Actor serves as the policy network \(\pi_\theta\), and the Graph Transformer Critic serves as the value network \(V_\phi\). The standard GAE advantage calculation used in baseline MAPPO is replaced by our counterfactual advantage, \(\tilde{A}^{CF}\). This agent-specific advantage is then used to update the shared actor network \(\pi_\theta\) via the clipped surrogate objective \eqref{eq:clip_loss_cf}. The centralized critic \(V_\phi\) is trained concurrently by minimizing the Huber loss or mean-squared error between its value predictions and the bootstrapped returns \(R_t\) over a set of trajectories \(\mathcal{D}\). The high-level training loop is presented in \Cref{alg:stacca}.

\begin{algorithm}[H]
\caption{STACCA w/ MAPPO Training Loop}
\label{alg:stacca}
\begin{algorithmic}[1]
\State Initialize Graph Transformer Actor \(\pi_\theta\) and Critic \(V_\phi\).
\For{each training iteration}
    \State \parbox[t]{0.9\linewidth}{Collect trajectories \(\mathcal{D}\) and one-step counterfactual states \(\mathcal{C}\) with policy \(\pi_\theta\) for all agents.}
    \State \parbox[t]{0.9\linewidth}{Compute state values for \(\mathcal{D}\) and \(\mathcal{C}\) with \(V_\phi\).}
    \State \parbox[t]{0.9\linewidth}{Compute bootstrapped returns using Generalized Advantage Estimation, \(R^{\tau}_t\), $\forall \tau,t$.}
    \State \parbox[t]{0.9\linewidth}{Compute counterfactual advantages \(\tilde{A}^{\tau,CF}_{i,t}\), $\forall \tau,i,t$ using \eqref{eq:cf_adv} and \eqref{eq:cf_norm}.}
    \For{\(K_{\pi}\) epochs}
        \State \parbox[t]{0.85\linewidth}{Update \(\pi_\theta\) by minimizing $L_{\pi}^{CLIP,CF}(\theta)$ on \(\mathcal{D}\).}
    \EndFor
    \For{\(K_{V}\) epochs}
        \State Update \(V_\phi\) by minimizing \(L_{V}(\phi)\) on \(\mathcal{D}\).
    \EndFor
\EndFor
\end{algorithmic}
\end{algorithm}

\section{Experiments and Analyses}
Our experiments are designed to answer several key questions regarding the framework's actor and critic architectures, credit assignment mechanism, and overall effectiveness. \textbf{(1)~Long-Range Interactions:} Do the long-range interactions captured by the critic's global self-attention and the structural inductive biases learned by GAT layers lead to better policies? \textbf{(2)~Network Generalizability:} Does STACCA's shared actor architecture improve policy learning and successfully generalize to network topologies and scales unseen during training? \textbf{(3)~Credit Assignment:} Does our proposed counterfactual advantage effectively address the multi-agent credit assignment problem, leading to improved and more stable learning for decentralized policies?

To answer these questions, we begin by describing the training environments. We then present a comprehensive set of ablation studies to validate the individual contributions of STACCA's core components. Finally, we conduct extensive generalization and scalability experiments to evaluate STACCA's ability to transfer policies to networks of varying structures and sizes. Our results demonstrate that STACCA not only outperforms baseline architectures but also learns robust, generalizable policies that are effective even under significant changes in the network environment.
\begin{figure}[!b]
    \centering
    \includegraphics[width=\linewidth]{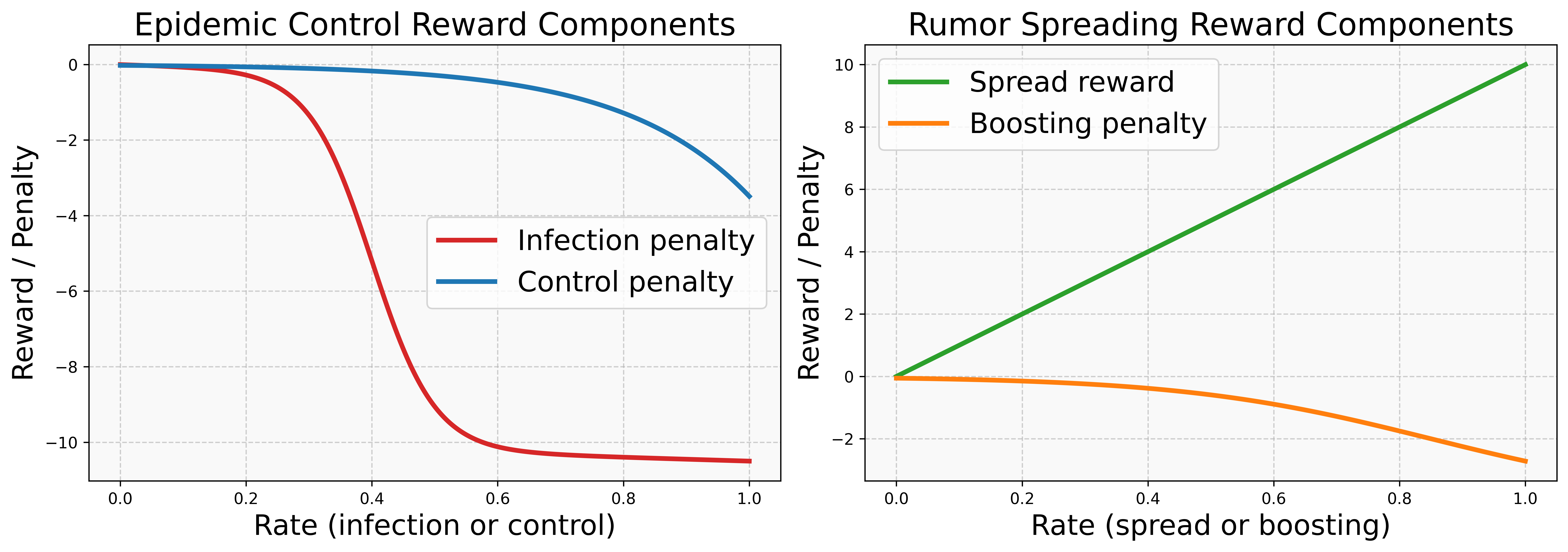}
    \caption{Reward Shaping Visualization.}
    \label{fig:reward_shaping}
\end{figure}

\begin{figure*}[!ht]
    \centering
    \begin{subfigure}[t]{0.325\textwidth}
        \centering
        \includegraphics[width=\linewidth]{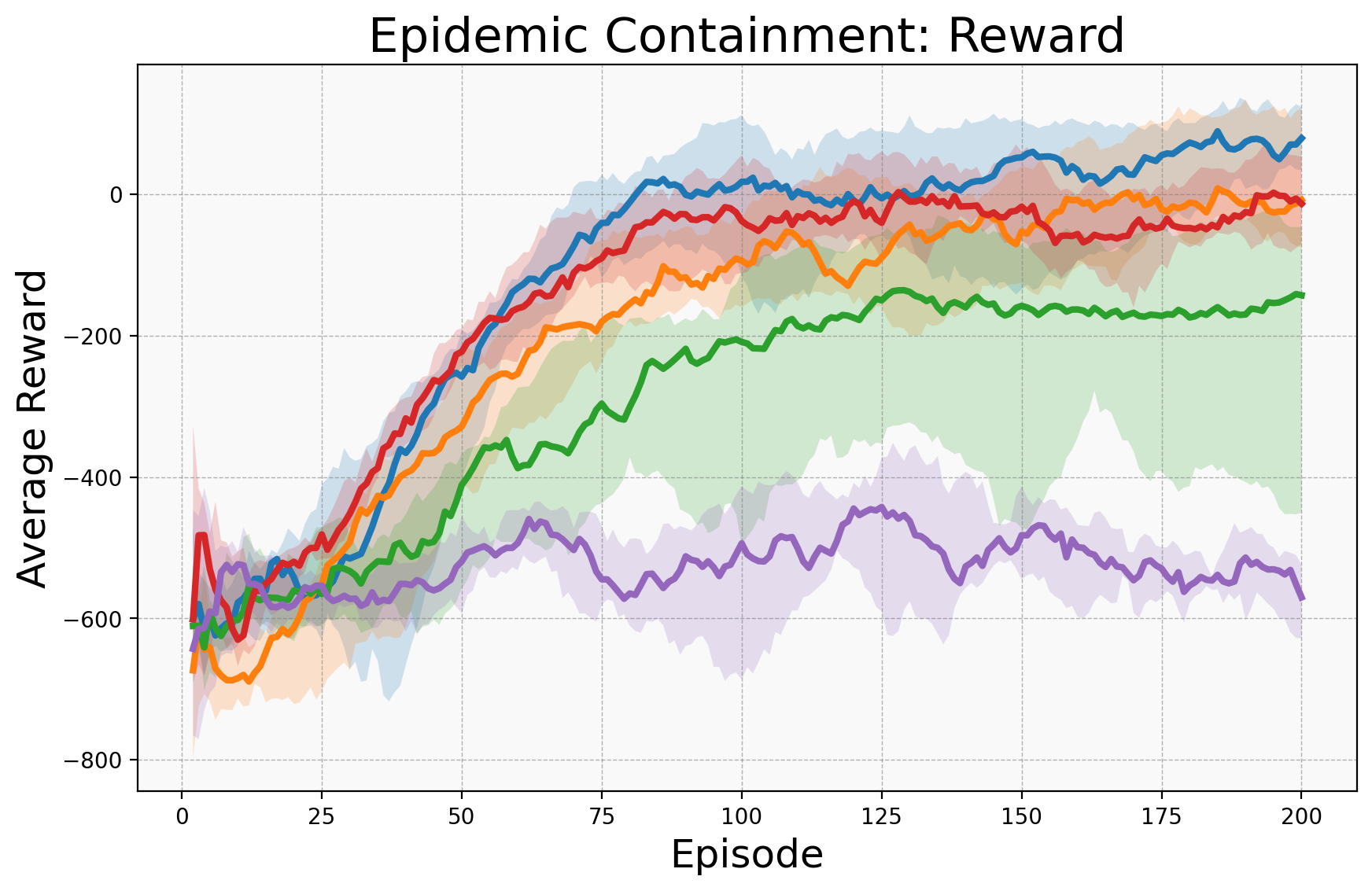}
    \end{subfigure}
    \hfill
    \begin{subfigure}[t]{0.325\textwidth}
        \centering
        \includegraphics[width=\linewidth]{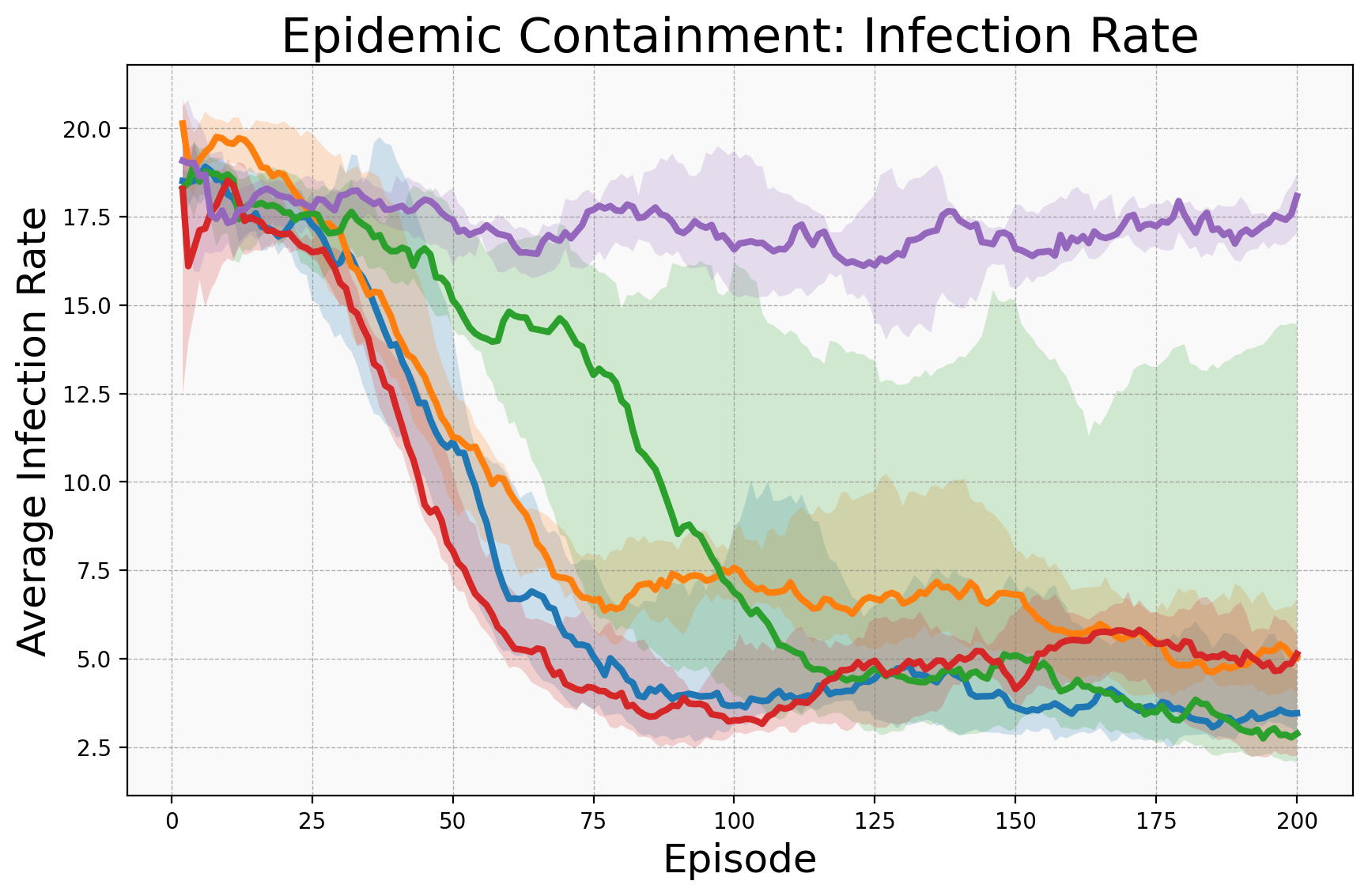}
    \end{subfigure}
    \hfill
    \begin{subfigure}[t]{0.325\textwidth}
        \centering
        \includegraphics[width=\linewidth]{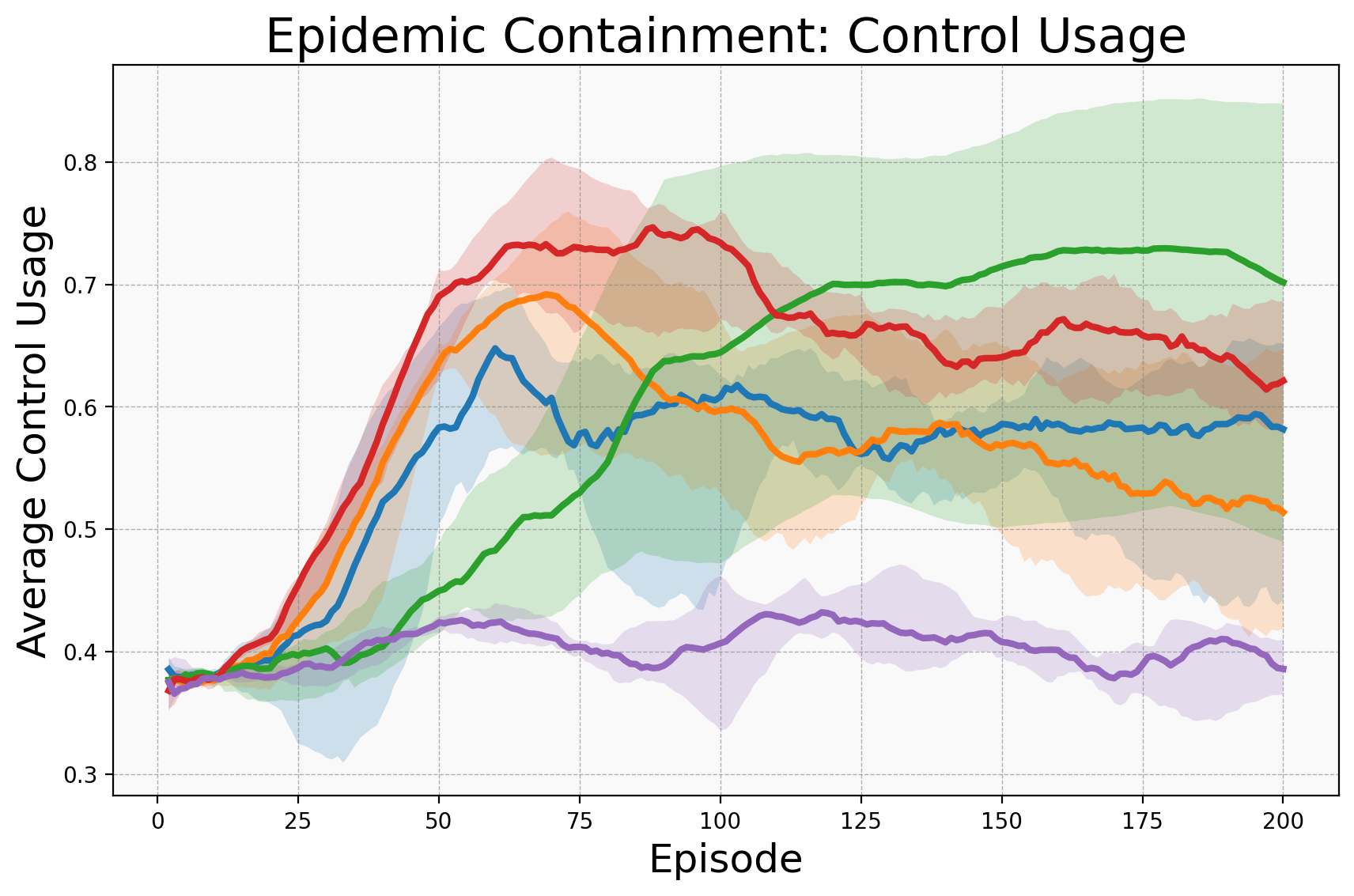}
    \end{subfigure}
    
    \begin{subfigure}[t]{0.325\textwidth}
        \centering
        \includegraphics[width=\linewidth]{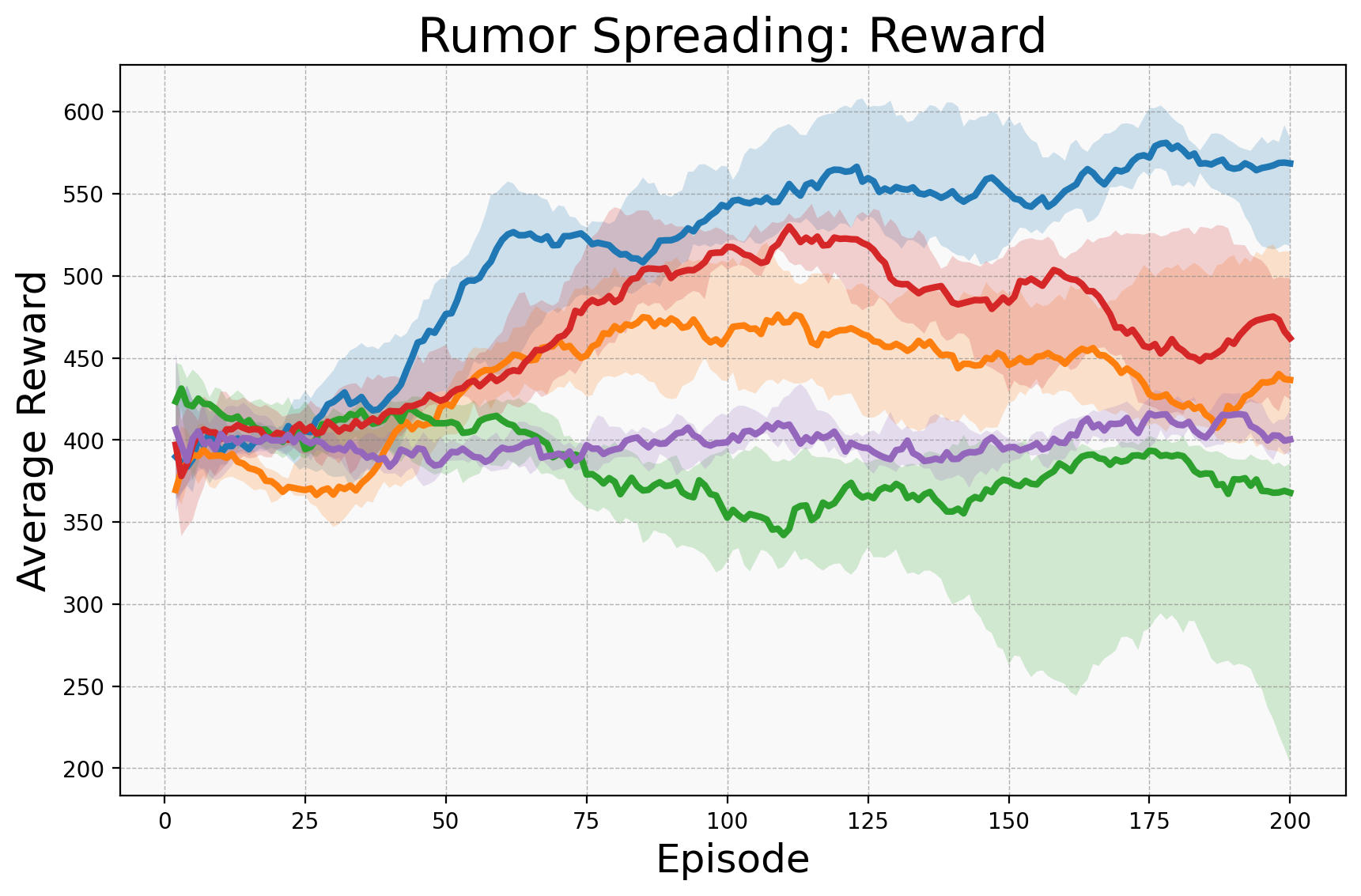}
    \end{subfigure}
    \hfill
    \begin{subfigure}[t]{0.325\textwidth}
        \centering
        \includegraphics[width=\linewidth]{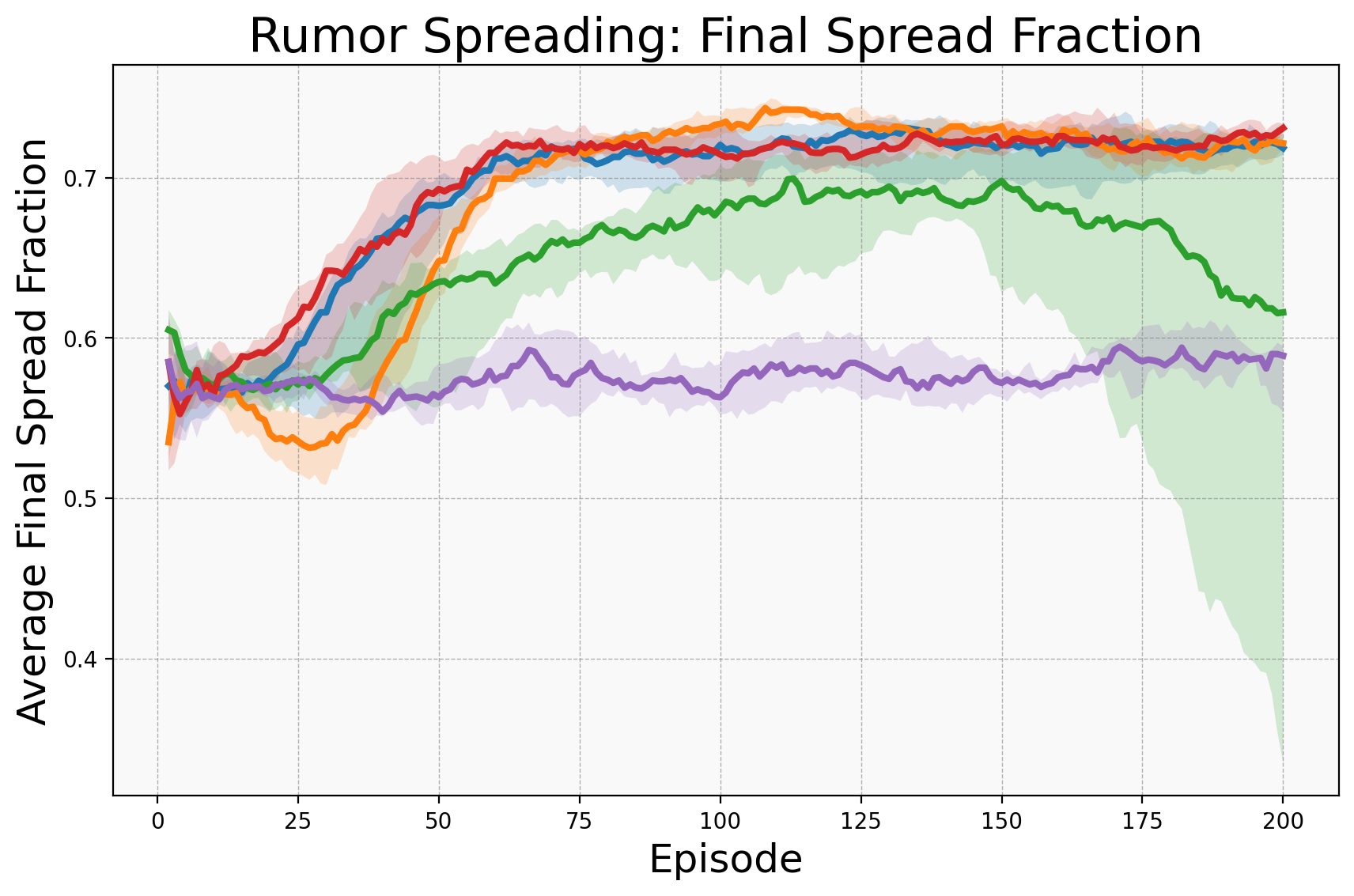}
    \end{subfigure}
    \hfill
    \begin{subfigure}[t]{0.325\textwidth}
        \centering
        \includegraphics[width=\linewidth]{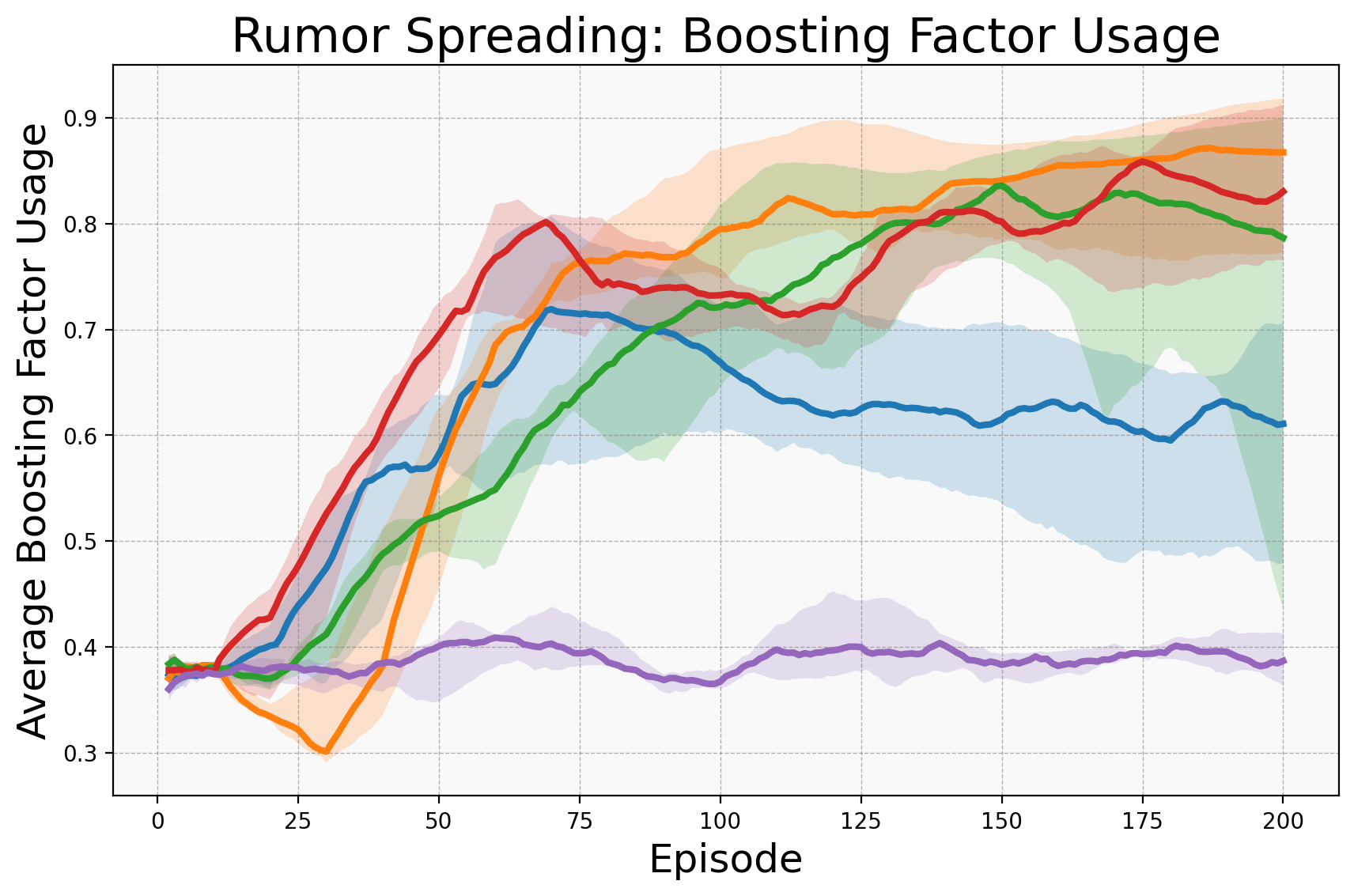}
    \end{subfigure}
    
    \vspace{0.0em}
    \includegraphics[width=0.8\textwidth]{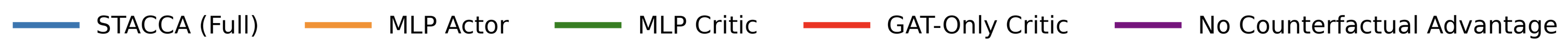}

    \caption{Ablation Experiments for STACCA: Comparing STACCA to STACCA w/ MLP Actor, STACCA w/ MLP Critic, STACCA w/ GAT-Only Critic, and STACCA w/ no counterfactual advantage for the epidemic containment environment (top row) and rumor-spreading environment (bottom row).}
    \label{fig:training}
\end{figure*}

\subsection{Training Environments}
For all training experiments, we use a 50-node Barabási-Albert (BA) \cite{barabasi1999emergence} graph with the attachment parameter $m=1$. This topology was chosen for its diverse degree distribution, which features a few high-degree hubs and many low-degree nodes. This heterogeneity exposes the actor to a wide variety of local neighborhood structures during training. The BA ($m=1$) graph also maintains a relatively high diameter, creating scenarios where the critic must learn long-range credit assignment to inform the actor's policy. Furthermore, applying \Cref{prop:long_range} from \Cref{sec:theory_mean_field} with the epidemic environment parameters used in our experiments ($\beta_0=0.25$ and $\delta=0.15$), this graph's linearized mean-field dynamics have a spectral radius of $\rho(J)=1.71$ and 14 spectral modes greater than 1, indicating high vulnerability to long-range interactions.

In both environments, each agent observes only its 1-hop local neighborhood. The action space is discrete, allowing agents to either increase, decrease, or maintain their current control (or boosting-factor) usage by increments of 0.1, with values clipped to the range $[0, 1]$. In the actor input, each node/agent is composed of five features: (1) a binary indicator identifying the ego node, (2) the infected/susceptible status in the epidemic setting or aware/unaware status in the rumor-spreading setting, (3) the current control or boosting-factor level, (4) the node degree, and (5) the shortest-path distance from the ego node. The critic input includes the corresponding node-level features for all nodes, excluding the ego-node indicator and distance features, since these are defined relative to each agent’s local perspective.

\subsubsection{Epidemic Containment Environment}
In this environment, agents apply control to limit the spread of an infection. Each training episode begins with three randomly selected seed nodes, a base transmission rate of $\beta_0=0.25$, and a control effectiveness of $\eta=0.9$. The reward function is designed to balance infection mitigation with cost-efficiency. An exponential penalty is applied to control usage, encouraging initial control usage, but discouraging excessive intervention. A softened catastrophe penalty, implemented as a flipped sigmoid function, penalizes the system when the number of infected nodes approaches an epidemic threshold. This is supplemented by a smaller linear penalty on the total number of infected nodes. This reward structure incentivizes agents to use control judiciously while preventing widespread outbreaks (see \Cref{fig:reward_shaping}). Additionally, a bonus of 3 per timestep is awarded when the infection is \emph{eradicated} (0 infected nodes), incentivizing rapid and complete elimination.

\subsubsection{Rumor Spreading Environment}
In this environment, the goal is to maximize the spread of information from a set of seed nodes (3 are used for training). We use a base transmission rate of $\beta_0=0.25$ (at full boosting factor) and a market saturation factor of $\kappa=3$, which makes spreading more difficult as more nodes become aware. For the reward function, an exponential penalty is applied to boosting-factor usage and a linear reward is applied to the aware nodes, directly incentivizing propagation.

\subsection{STACCA Ablation Experiments}
We conduct a series of ablation studies to isolate and quantify the contribution of each key component of STACCA (see \Cref{fig:training}). To evaluate the critic architecture, we compare the STACCA critic to a version with the global attention mechanism removed (GAT-only) and a version with both graph-based and global attention removed (MLP architecture). To evaluate the actor architecture, we compare the STACCA actor to an MLP actor. Finally, we compare STACCA's counterfactual advantage to the standard GAE with trajectory-level normalization, as is conventionally performed in MAPPO.

\subsubsection{Critic Comparison}
Incorporating the graph-based and global attention mechanisms demonstrated a clear performance benefit. The MLP critic performed sub-optimally, highlighting the value of the structural inductive bias from GAT layers. Adding the global transformer on top of the GAT layers yielded further improvements, with a more pronounced effect in the rumor-spreading environment. This suggests that the global attention mechanism is particularly vital when task dynamics, such as the saturation component in rumor spreading, depend on global state information.

\subsubsection{Actor Comparison} The STACCA actor outperformed the MLP actor in both environments, with a more significant improvement in the rumor-spreading task. However, the primary benefit of the STACCA actor architecture lies in its generalizability, a crucial feature evaluated in \Cref{subsec:network_gen}.

\subsubsection{Advantage Comparison} The counterfactual advantage had a significant impact on performance in both scenarios. The conventional GAE, applied uniformly to all agents, provides a weak and noisy learning signal in a system with 50 agents. This result underscores the importance of agent-specific credit assignment in large-scale MARL settings.

\begin{figure*}[!ht]
    \centering
    \begin{subfigure}[t]{0.24\textwidth}
        \centering
        \includegraphics[width=\linewidth]{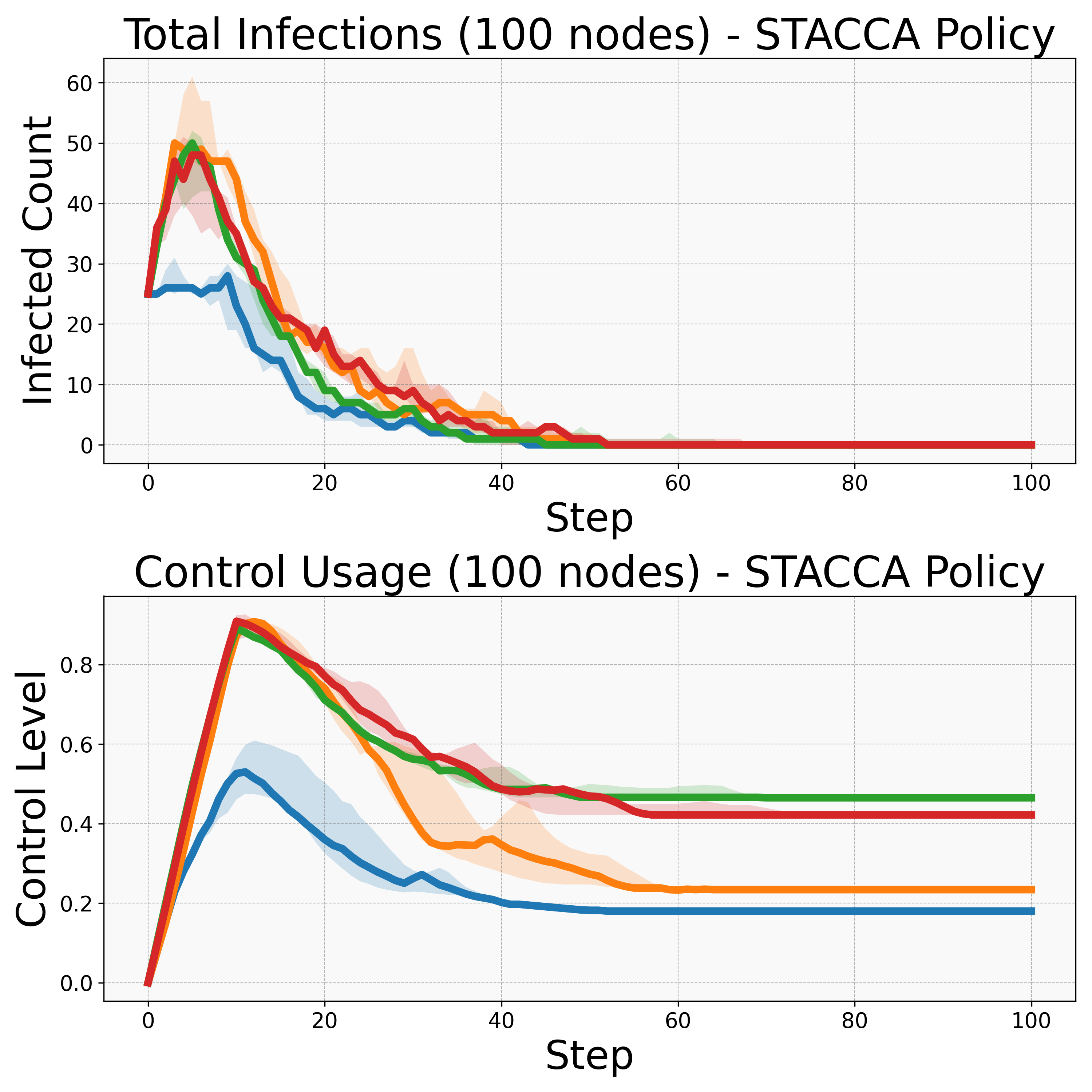}
    \end{subfigure}
    \hfill
    \begin{subfigure}[t]{0.24\textwidth}
        \centering
        \includegraphics[width=\linewidth]{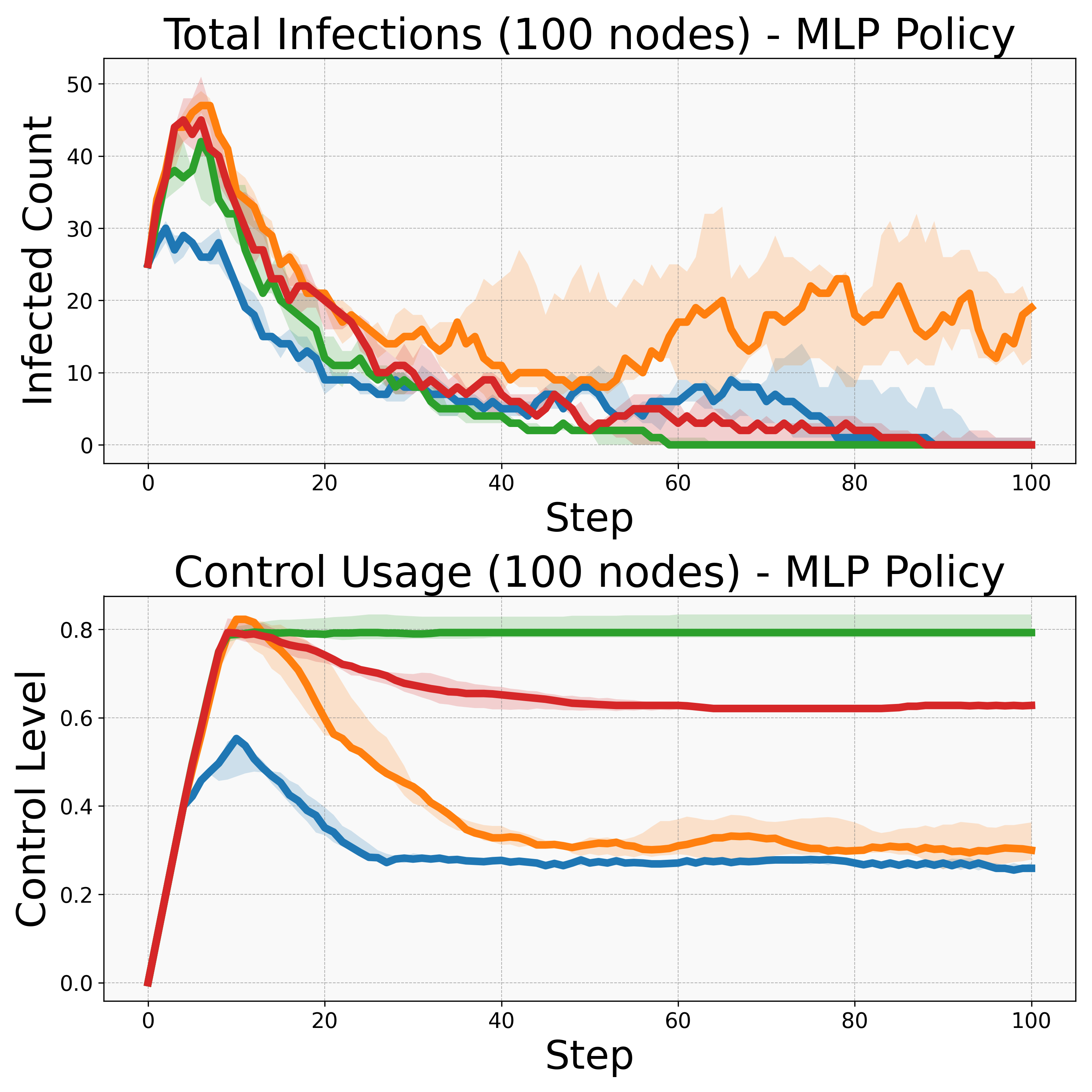}
    \end{subfigure}
    \hfill
    \begin{subfigure}[t]{0.24\textwidth}
        \centering
        \includegraphics[width=\linewidth]{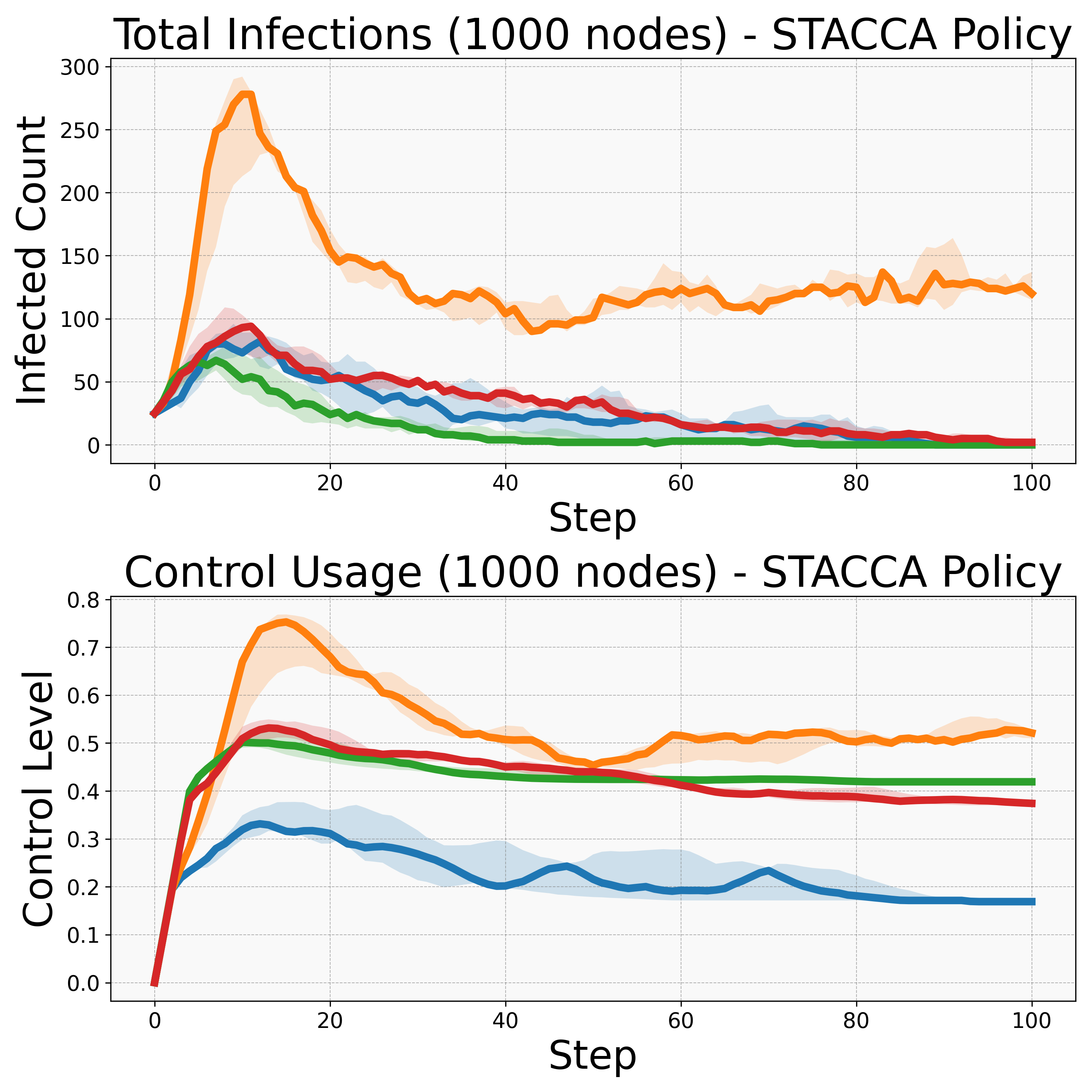}
    \end{subfigure}
    \hfill
    \begin{subfigure}[t]{0.24\textwidth}
        \centering
        \includegraphics[width=\linewidth]{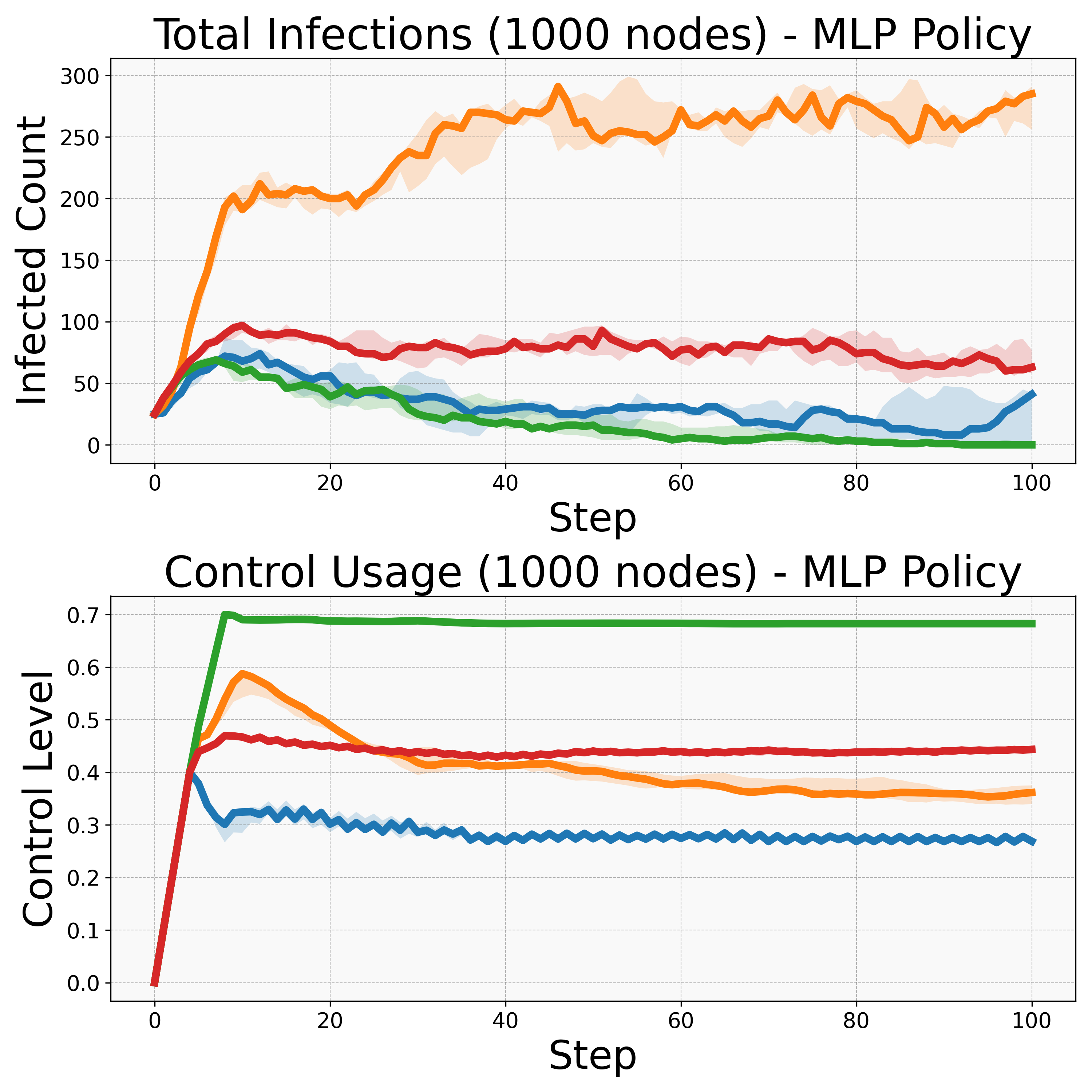}
    \end{subfigure}

    \vspace{0em}
    \includegraphics[width=0.9\textwidth]{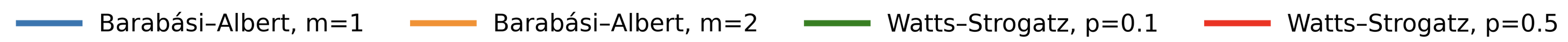}

    \caption{Epidemic Containment Environment: Comparison of MLP Policy and STACCA Graph Transformer Policy on 100-node and 1000-node graphs of each of the 4 graph types. All examples use 25 infected seed nodes.}
    \label{fig:epidemic_comparison}
\end{figure*}

\subsection{Network Generalization and Scalability Experiments}\label{subsec:network_gen}

To evaluate the ability of STACCA's learned policies to generalize, we test policies trained on the 50-node BA ($m=1$) graph on four unseen network topologies at two different scales (100 and 1000 nodes). The test graphs include two Watts-Strogatz (WS) graphs ($k=4$) \cite{watts1998collective} with rewiring probabilities $p=0.1$ and $p=0.5$, and two Barabási-Albert (BA) graphs with $m=1$ and $m=2$.

We compare the performance of the STACCA actor against an identically trained MLP actor. While prior networked MARL frameworks (e.g., \cite{qu2020scalable, qu2020scalable_local}) introduce valuable structural biases, their architectures are predominantly designed and tuned for fixed-topology training. Adapting them for zero-shot topological generalization would require significant architectural modifications, confounding the comparison. Thus, we use an MLP actor baseline to isolate the impact of STACCA's actor architecture on network generalizability. For each actor type, we use the best-performing model based on rewards from 10 independent training runs.

\subsubsection{Epidemic Containment}
As shown in \Cref{fig:epidemic_comparison}, the STACCA policy demonstrates strong generalization. For the 100-node graphs, the STACCA actor successfully eradicates the infection on all four graph types within 50 timesteps using moderate control. In contrast, the MLP actor takes nearly twice as long on three of the graphs with higher control usage, and fails to eradicate the infection on the BA ($m=2$) graph, where $\sim$20\% of nodes remain infected. At a larger scale (1000 nodes), the STACCA actor still eradicates the infection on three of the four graphs. On the more challenging BA ($m=2$) graph, it contains the infection to 10--15\% of the population. The MLP actor's performance degrades substantially at this scale, failing to eradicate the infection on three of the four graphs and allowing severe outbreaks ($\sim$30\% infected) on the BA ($m=2$) graph.

The reduced performance on the BA ($m=2$) graph, especially at scale, can be attributed to its degree distribution. As BA graphs grow, the degree of hub nodes increases significantly, creating dense local neighborhoods unseen during training. This effect is more pronounced for $m=2$, where the total edge density is approximately double that of $m=1$. This combination of out-of-distribution hub nodes and higher graph connectivity exacerbates the epidemic dynamics (see \Cref{epidemic_dynamics}), leading to higher and less controllable infection rates. The more uniform local structure of WS graphs likely contributes to the more consistent performance.

\begin{figure}[!ht]
    \centering
    \begin{subfigure}[c]{0.71\linewidth}
        \centering
        \includegraphics[width=\linewidth]{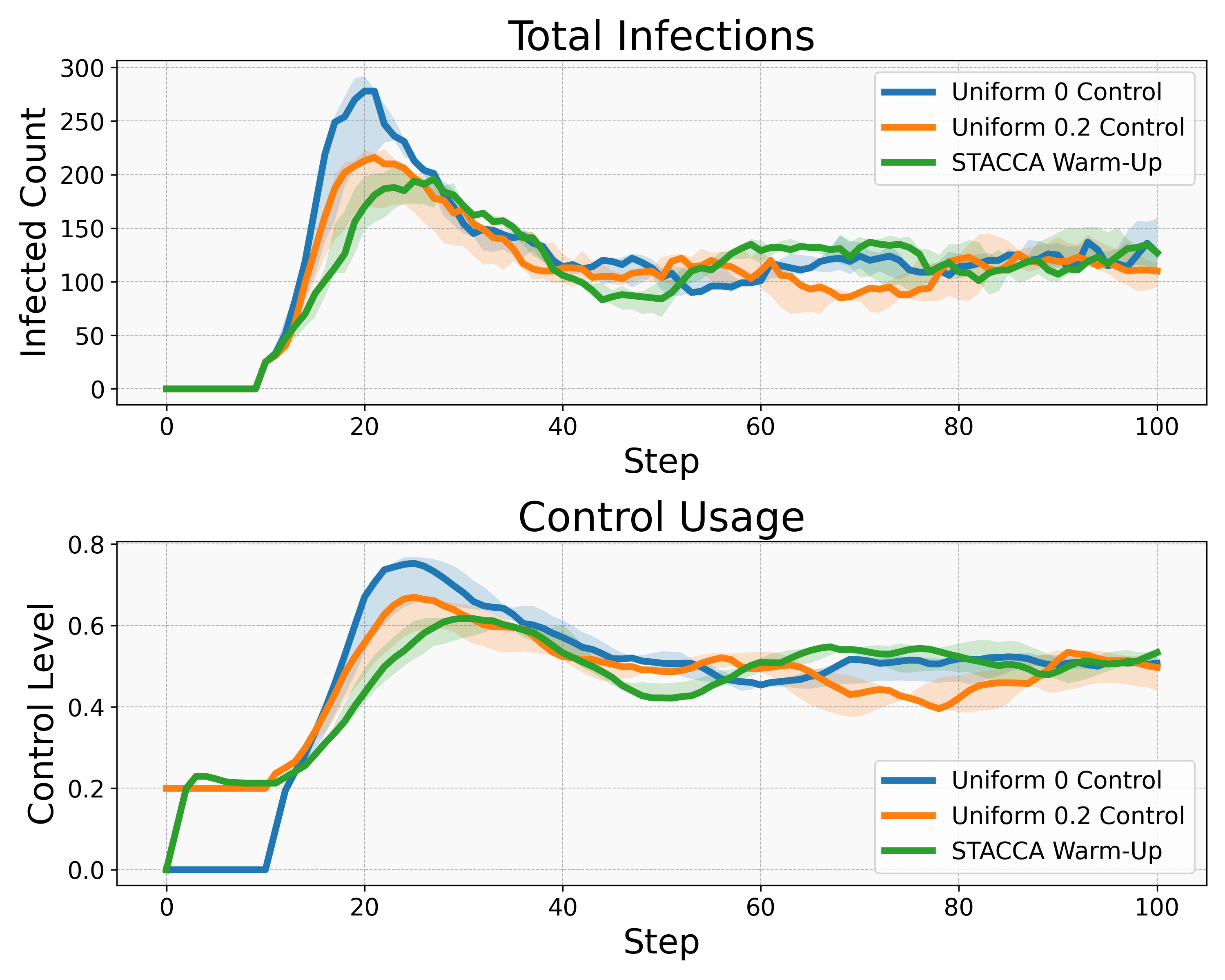}
    \end{subfigure}
    \hfill
    \begin{subfigure}[c]{0.27\linewidth}
        \centering
        \includegraphics[width=\linewidth]{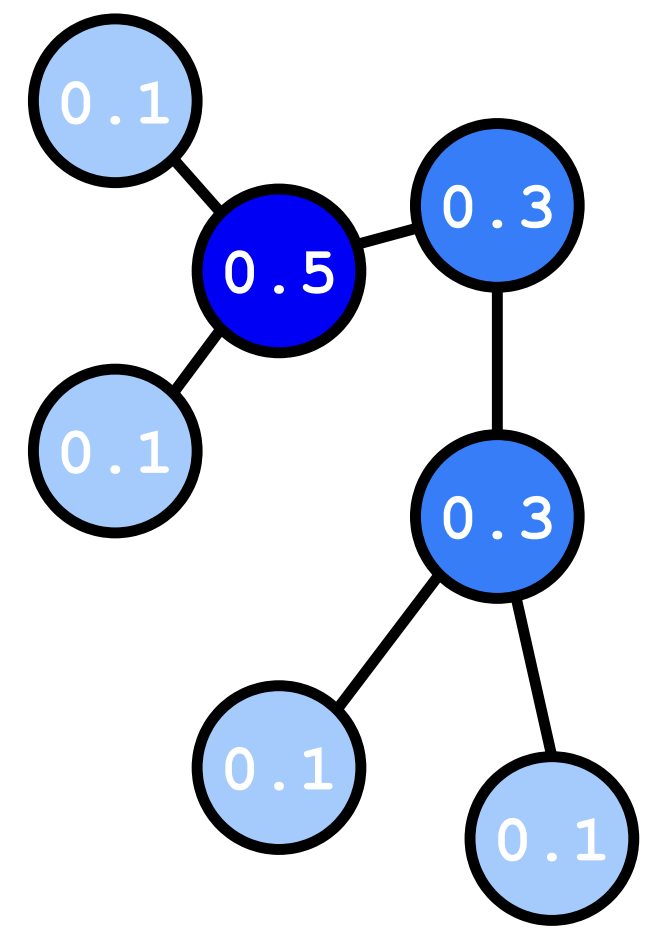}
    \end{subfigure}
    \caption{Node initialization comparison on BA ($m=2$) 1000-node graph. 25 infected seeds are introduced at $t=10$ (left). Visual toy example of a precautionary baseline state (right).}
    \label{fig:emergent}
\end{figure}

We also observe an emergent behavior where the STACCA policy maintains a low level of control even after eradication. This behavior reduces the peak number of infections in subsequent outbreaks. As shown in \Cref{fig:emergent}, this precautionary state is more effective than a uniform control initialization of equivalent global control usage. However, the difference is minor, so we leave this for further investigation.

\subsubsection{Rumor Spreading}
\begin{figure}[!ht]
    \centering
    \includegraphics[width=\linewidth]{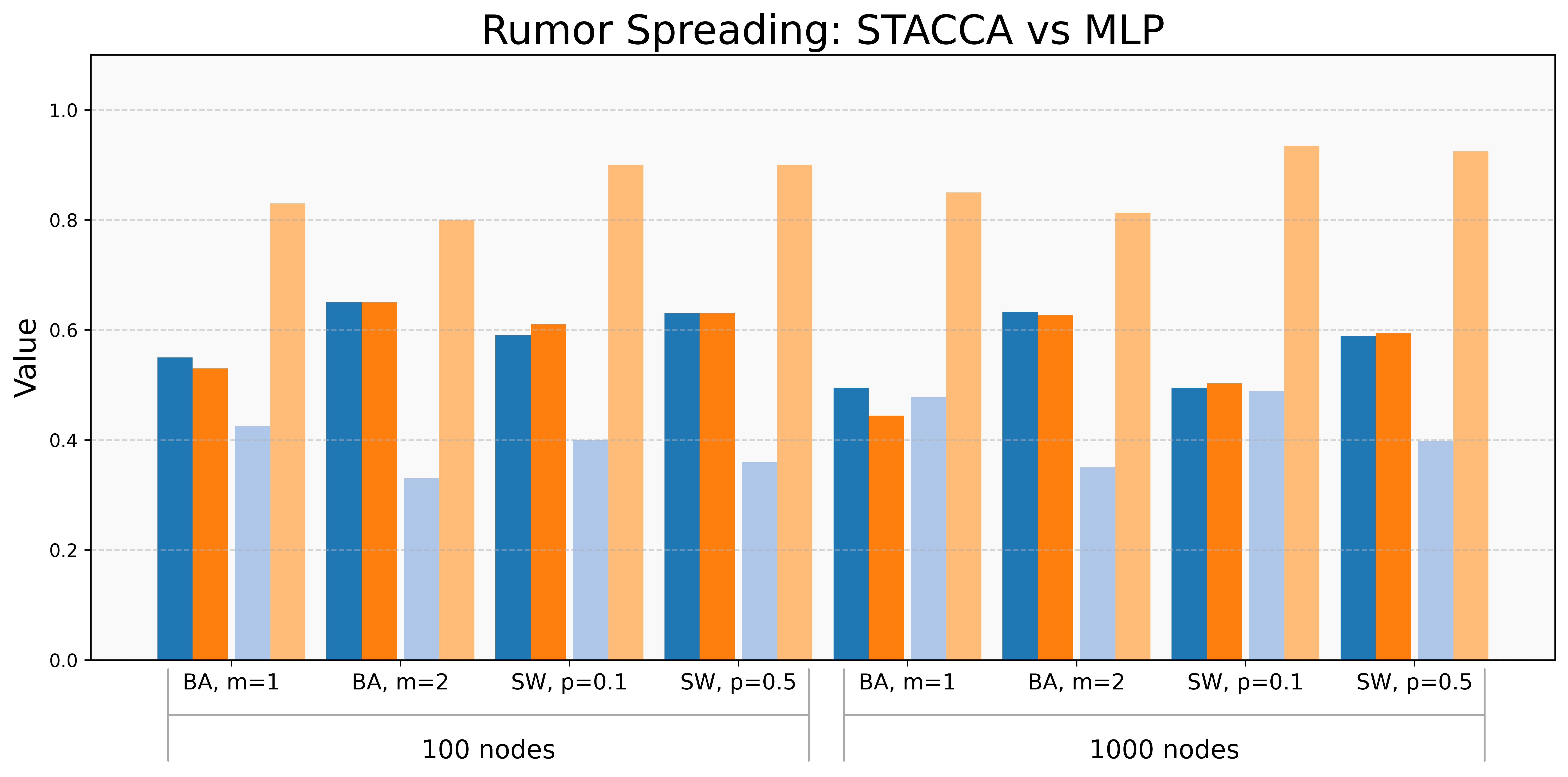}
    \vspace{0.0em}
    \includegraphics[width=1.0\linewidth]{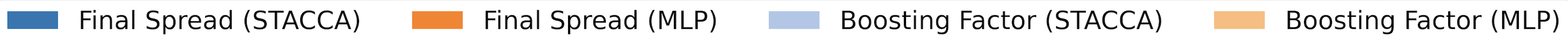}
    \caption{Rumor Spreading Environment: STACCA Policy vs. MLP Policy on final spread proportion and average boosting-factor usage over 100 timesteps (5 seed nodes).}
    \label{fig:rumor_scale}
\end{figure}

As shown in \Cref{fig:rumor_scale}, the STACCA actor generalizes effectively across different network structures. It achieves higher final spread on denser networks (BA ($m=2$), WS ($p=0.5$)), effectively exploiting the network structure. The MLP actor achieves a comparable final spread but with excessive boosting-factor usage. At 1000 nodes, STACCA maintains its efficient, high-performance strategy. While the final spread fraction is slightly lower, this is likely an artifact of the fixed 100-timestep evaluation window. The MLP policy continues to exhibit high-cost, inefficient behavior at scale.

\section{Conclusion}
In this work, we first studied long-range interactions in networked systems from a mean-field stability perspective. We analyzed how these approximate dynamics and their stability conditions translate to the true system through an empirical study. This analysis showed that the linearized mean-field system reveals strong correlations with long-range interactions, providing motivation for the design choices behind our primary contribution: STACCA. We aimed to address two key challenges in the application of MARL solutions to networked systems: (1) capturing and exploiting long-range network interactions and (2) generalizing across network topologies and scales. STACCA's Graph Transformer Critic learns global, long-range system dynamics, while its shared Graph Transformer Actor learns a network-generalizable policy, enabling zero-shot transfer to unseen networks. By integrating a novel counterfactual baseline into the advantage formulation, STACCA addresses the credit assignment problem at scale. Our experiments show that STACCA successfully generalizes to networks of diverse topological structures and significantly larger scales than those seen during training. These results highlight the potential of transformer-based architectures to create scalable, adaptable, and effective control policies for complex, real-world multi-agent systems.

\addtolength{\textheight}{-12cm}   % This command serves to balance the column lengths

%%%%%%%%%%%%%%%%%%%%%%%%%%%%%%%%%%%%%%%%%%%%%%%%%%%%%%%%%%%%%%%%%%%%%%%%%%%%%%%%

{
\small
\bibliographystyle{plain}
\bibliography{main}
}

\end{document}